# Optimizing Predictive Maintenance in Intelligent Manufacturing: An Integrated FNO-DAE-GNN-PPO MDP Framework

Shiqing Qiu[1]

1. School of mathematical sciences, Chengdu University of Technology, China

**Abstract**

In the era of smart manufacturing, predictive maintenance (PdM) plays a pivotal role in improving equipment reliability and reducing operating costs. In this paper, we propose a novel Markov Decision Process (MDP) framework that integrates advanced soft computing techniques - Fourier Neural Operator (FNO), Denoising Autoencoder (DAE), Graph Neural Network (GNN), and Proximal Policy Optimisation (PPO) - to address the multidimensional challenges of predictive maintenance in complex manufacturing systems. Specifically, the proposed framework innovatively combines the powerful frequency-domain representation capability of FNOs to capture high-dimensional temporal patterns; DAEs to achieve robust, noise-resistant latent state embedding from complex non-Gaussian sensor data; and GNNs to accurately represent inter-device dependencies for coordinated system-wide maintenance decisions. Furthermore, by exploiting PPO, the framework ensures stable and efficient optimisation of long-term maintenance strategies to effectively handle uncertainty and non-stationary dynamics. Experimental validation demonstrates that the approach significantly outperforms multiple deep learning baseline models with up to 13% cost reduction, as well as strong convergence and inter-module synergy. The framework has considerable industrial potential to effectively reduce downtime and operating expenses through data-driven strategies.

## 1. Introduction

As science and technology drive the transformation of manufacturing toward intelligent manufacturing, enterprise production has become highly dependent on automated and precision-engineered machine tool assembly lines [1]. In an intelligent manufacturing environment, the health status and operational efficiency of equipment directly determine the stability of production processes and the final product quality pass rate [2-3]. To ensure continuous, high-quality production and meet increasingly stringent market expectations, proactive and efficient maintenance of production equipment has become critical [4]. Traditional scheduled maintenance or reactive repair strategies based on equipment failures often struggle to effectively predict and prevent issues in complex systems and dynamic operating conditions, leading to potential high downtime costs and

───────────────
*Corresponding author at:

Email addresses: qiusq04@163.com(S. Qiu)

production losses [5]. Therefore, developing advanced predictive maintenance strategies that leverage real-time data and intelligent algorithms to optimize maintenance decisions, thereby minimizing fault risks, reducing total maintenance costs, and ensuring optimal performance of production systems, has become a critical requirement for modern manufacturing enterprises to enhance their core competitiveness [6].

Formulating the aforementioned complex maintenance decision challenges into a sequential decision problem is a critical step in developing intelligent solutions. Markov Decision Processes (MDPs) provide a robust mathematical foundation for sequential decision problems in uncertain environments and have been applied in various fields, including equipment maintenance [7]. However, when directly applying traditional MDP frameworks to complex equipment maintenance scheduling in the context of intelligent manufacturing, significant challenges arise. In El Chamie's paper, they introduced a novel sensor measurement in the MDP model that provided additional information about the stochastic process, which can therefore be incorporated into decision-making strategies to improve performance [8]. The studies of Anantharam, Qiu,S., and Ying showed that the policy iteration algorithm for finding the optimal policy can be greatly simplified in all discrete-time reversible Markov decision problems with finite state and action spaces [9-11]. On the basis of a continuous-time Markov decision process, Piunovskiy et al. and Golui et al. efficiently established a linear programming method for solving related constrained optimal control problems under rather natural conditions [12,13], but the engineering landability was still limited and no scalable approximation strategy could be given. Therefore, investigating how to integrate advanced machine learning techniques (such as deep reinforcement learning, representation learning, and graph neural networks) to overcome the limitations of traditional MDPs and achieve efficient, adaptive optimization decisions for complex systems holds important theoretical and practical significance.

Although existing research has made progress in optimizing equipment maintenance, it still faces some key challenges. For example, Roy et al. proposed a structure-aware learning algorithm to exploit the ordered multi-threshold structure of the optimal policy, but it only applied to finite-state-action, fully sizable CTMDPs [14]. In Savas's paper, they provided an exact algorithm based on mixed-integer linear programming for making optimal decisions [15]. However, there was still much room for improvement in complex constraint handling and empirical evaluation. Chelouati et al. assumed that the decision maker did not know the state before making a decision, and according to Markov distribution evolution, they could obtain practical solutions even in high-dimensional problems [16,17], but the method lacked the robustness and generalisation. Many methods struggle with the inherent challenges of handling the high-dimensional and strongly time-series-correlated nature of industrial sensor data; there are limitations in constructing high-fidelity, robust state representations that can capture the intrinsic features of complex (e.g., non-Gaussian or multimodal) equipment states; and the significant impact of inter-device dependencies (e.g., production line constraints or resource competition) on global optimal maintenance strategies is often overlooked. Furthermore, integrating these elements into an end-to-end framework capable of stably and efficiently learning long-term optimal maintenance strategies remains an area that requires further exploration.

To address the multidimensional, dynamic, and structured challenges of predictive maintenance in complex manufacturing systems, this paper proposes a Markov Decision Process (MDP) framework that integrates multiple soft computing techniques to optimize equipment maintenance strategies and reduce long-term costs and downtime risks. First, to address the high-dimensional temporal characteristics of equipment sensor data, this paper uses Fourier Neural Operators (FNO) for frequency domain modeling to capture complex dynamic patterns in equipment state evolution. To extract more robust and compact state representations, the Denoising Autoencoder (DAE) is introduced to perform nonlinear dimensionality reduction on local sensor features, enhancing the model's adaptability to noise and non-Gaussian feature distributions. Furthermore, considering the interactive influences between devices in production systems, a Graph Neural Network (GNN) based on device group structures is constructed to integrate the states of neighboring devices, enabling the modeling of structured contextual information. Finally, based on the high-quality state embeddings and graph structure representations, the Proximal Policy Optimization (PPO) algorithm is used to train and optimize maintenance strategies. PPO achieves optimal control of long-term maintenance costs by fully utilizing historical experience with a stable and efficient policy gradient method. Through this multi-module collaborative mechanism, this study systematically addresses the core challenges of predictive maintenance from time-series modeling to state representation and strategy optimization.

The primary contributions of this work are outlined as follows:

(1) A hybrid temporal-frequency modeling approach is introduced, utilizing FNO to capture long-range temporal dependencies and frequency-domain dynamics in high-dimensional sensor data, supported by a DAE for noise-resilient representation learning.

(2) We embed device interdependencies through GNN, allowing structured information propagation across equipment groups, enabling context-aware and system-level maintenance decision optimization.

(3) The PPO algorithm is applied for policy optimization, offering a stable, sample-efficient reinforcement learning method tailored to minimize long-term operational costs under uncertainty.

(4) We propose an integrated decision-making framework for predictive maintenance that combines FNO, DAE, GNN, and PPO, formulated within a MDP structure to address complex industrial environments.

(5) Comprehensive experiments and ablation studies on synthetic industrial data demonstrate the proposed framework's superior cost-efficiency, stability, and scalability compared to multiple baselines, validating the contribution of each individual module.

The remainder of this paper is organized as follows. Section 2 reviews the research work in the field of equipment maintenance optimization and related technologies. Section 3 summaries the problem statement and the preliminaries. In Section 4, it elaborates on the methodology, detailing the proposed MDP framework based on FNO-DAE-GNN-PPO. Section 5 presents the experimental results and performance evaluation, covering comparisons with baseline models, computational complexity analysis, convergence verification, and the analyses in real industry datas. Section 6 analyzes the findings, discusses their potential in industrial applications, addresses the limitations of the current method, and outlines future research directions. Finally, Section 7 concludes the paper.

## 2. Related work

2.1. Predictive maintenance and MDP formulation

Maintenance strategies for industrial equipment have evolved from a reactive approach of repairing failures after they occur and conducting fixed-schedule inspections to a more proactive and intelligent paradigm. Traditional strategies often come with high unexpected downtime costs or unnecessary waste of maintenance resources. Predictive maintenance (PdM) has emerged as an advanced maintenance concept, focusing on analyzing real-time monitoring data to assess equipment health, predict the likelihood of future failures, and determine the optimal timing and measures for maintenance interventions. The goal is to maximize equipment availability and minimize total maintenance costs [18].

Given that equipment maintenance decisions need to consider long-term cumulative effects under uncertainty (such as degradation rates and repair effectiveness), MDP provides a powerful theoretical framework for this problem. A typical maintenance MDP model defines equipment health status as the state space and possible maintenance interventions (such as continue operation, inspection, repair, or replacement) as the action space [19,20]. The transition probabilities between states capture the random degradation and repair dynamics of the equipment under different maintenance strategies, while the reward or cost function quantifies the economic consequences of executing specific maintenance actions and the system being in specific states. By solving the MDP, it is theoretically possible to find an optimal maintenance strategy that specifies the best action to take in each state to maximize long-term expected returns or minimize long-term expected costs [21].

However, directly applying the classic MDP framework to real-world industrial maintenance scenarios poses significant challenges. The state space of complex equipment is often extremely high-dimensional or even continuous, leading to the so-called "dimension disaster" [22]. Additionally, precise state transition probabilities and cost functions are often difficult to obtain a priori or accurately estimate from limited data. Furthermore, the true internal health status of equipment is typically not fully observable, making the problem more aligned with the partially observable MDP (POMDP) framework [23]. These challenges limit the direct application of traditional MDP solving techniques, which rely on precise models, in the optimization of complex system maintenance. Therefore, there is an urgent need to develop advanced data-driven methods that can effectively utilize large amounts of available sensor data without requiring precise models, such as the deep learning and reinforcement learning techniques discussed in subsequent chapters.

2.2. Data-driven equipment state assessment and representation learning

To address these complex equipment maintenance challenges, existing research has explored various approaches from multiple angles. For example, degradation analysis methods based on physical models, such as the crack propagation model based on Paris's law [24,25] or the chemical aging model based on the Arrhenius equation [26,27], although they offer good interpretability, typically require precise prior knowledge of material properties, load conditions, and failure mechanisms. and constructing models applicable to the entire complex system rather than just a single simple component is often very challenging, making them difficult to adapt to the variable operating conditions and diverse equipment types in smart manufacturing. Traditional machine learning methods, such as support vector machines (SVM) [28,29] or random forests [30,31], have achieved some success in fault diagnosis and remaining life prediction, such as predicting the remaining life of bearings or gearboxes by extracting time-domain and frequency-domain features. However, these methods often require cumbersome and expert-dependent feature engineering and typically simplify maintenance problems into single-step classification or regression tasks, making it difficult to directly optimize sequential maintenance decisions that must consider long-term cumulative costs and benefits.

To overcome the limitations of feature engineering, deep learning (DL) methods, particularly recurrent neural networks (RNNs) and their variants (e.g., LSTM, GRU) and convolutional neural networks (CNNs), have been widely applied to learn features directly from simulation time-series data [32,33]. These models can automatically capture time-dependent and local patterns in data, demonstrating strong capabilities in equipment fault diagnosis and RUL prediction. For example, in Wahid's research [34], to capture abstract features of factory machine sensors and optimize prediction strategies, they proposed a predictive maintenance (PdM) multivariate time series prediction method combining convolutional neural networks and long short-term memory with skip-connections (CNN-LSTM). This method achieved the most reliable and highest prediction accuracy in the dataset provided in Wahid's paper [24]. However, these standard deep learning models may also face the following challenges. First, RNN-based models may suffer from gradient issues due to the potentially very long-term dependencies present in simulations. Second, standard CNN or RNN architectures may not be the most efficient at capturing specific global dependencies or frequency-domain characteristics present in simulation data. For example, if the simulation aims to model subtle changes in resonance phenomena or periodic behavior, these models may not be able to directly and effectively capture these critical frequency-domain information.

To effectively address the issue of critical frequency-domain information being overlooked, this study introduces the Fourier neural operator. The Fourier Neural Operator (FNO) is an emerging deep learning architecture designed to learn mappings between infinite-dimensional function spaces, particularly suited for solving scientific computation problems such as partial differential equations [35]. Unlike traditional neural networks that process finite-dimensional vectors, FNO directly operates on functions in the Fourier domain by parameterizing the integral kernels in the Fourier transform, enabling efficient and expressive learning. This method leverages the efficiency of the fast Fourier transform to effectively capture global dependencies and complex spatial patterns, demonstrating computational speeds several orders of magnitude faster than traditional numerical solvers while maintaining high accuracy in various physical simulation tasks [36].

When making maintenance decisions in an MDP or reinforcement learning framework, a core challenge is how to learn a low-dimensional but information-rich state representation from high-dimensional observation data. This representation needs to effectively capture key information about the true health status of the system to guide subsequent policy learning. Traditional methods such as linear dimension reduction techniques like PCA or non-linear Variational Autoencoders (VAE) compress input data into a latent space through unsupervised learning [37,38]. However, when simulation data aims to model complex, even multimodal, device state distributions or non-Gaussian noise processes, standard VAEs may tend to produce overly smooth reconstructions that struggle to accurately capture the details of these complex distributions, potentially losing information critical for distinguishing between key states [37]. Denoising autoencoders (DAEs) are a class of representation learning neural network models. Its core idea consists of two main parts: an encoder and a decoder [39]. During training, the model receives the original input data with added noise, and the encoder compresses it into a low-dimensional latent representation that captures the key features of the data [40]. The decoder then attempts to reconstruct the original, noise-free input data from this latent representation. By training the model to reconstruct the original signal in the presence of noise, DAE can learn noise-insensitive, more robust data features, making it highly suitable for extracting useful device status information from sensor data with interference or anomalies, and achieving effective dimensionality reduction.

In summary, existing data-driven methods have their own advantages and disadvantages in assessing equipment status and learning status representations from simulation data. Traditional methods rely on manual features and are difficult to deal with complex dynamics; standard deep learning models still have room for improvement in capturing specific temporal characteristics and maintaining the fidelity of status representations. Effectively dealing with complex temporal dynamics in simulated data and learning low-dimensional representations that accurately capture underlying state distributions are key to improving the performance of data-driven maintenance decisions. However, accurately assessing the state of a single device is often insufficient. The interactions between devices in a manufacturing system are equally critical for determining the optimal maintenance strategy at the system level.

2.3. Modeling interdependencies in manufacturing systems

Although the methods described in Section 2.2 are designed to improve the accuracy of individual device status assessment and the quality of status representation, devices do not operate in isolation in typical manufacturing environments. Devices on a production line are often interconnected through shared resources or control logic, forming a complex system. In such a system, the failure, performance degradation, or maintenance activities of a single device inevitably cause a chain reaction that affects the production efficiency of upstream and downstream devices and even the entire system. Therefore, maintenance decision optimization based solely on the status of individual devices may lead to locally optimal solutions but fail to achieve overall system-level performance optimization. Identifying and effectively modeling these interdependencies between devices is a critical step in developing globally optimal maintenance strategies.

Many studies simplify the problem by assuming that devices are independent of one another, which clearly does not align with the reality of most manufacturing systems. Some studies have attempted to capture the dynamic interactions between devices and system-level effects using system simulation techniques such as discrete event simulation (DES) or agent-based modeling (ABM) [41,42]. While these methods can simulate complex system behavior, constructing accurate simulation models can be highly complex and time-consuming, and they often struggle to integrate directly and efficiently with data-driven state assessment and online reinforcement learning decision-making frameworks. Additionally, these methods typically focus on simulating behavioral outcomes rather than explicitly learning or leveraging the system's intrinsic, structured dependency information.

Existing methods generally lack explicit modeling and utilization of the inherent network topology of manufacturing systems. Whether it is the physical layout of a production line, the upstream and downstream relationships in a process flow, or a cluster of devices sharing the same critical resource, these can all be naturally abstracted into a graph, where nodes represent devices and edges represent the dependencies between them [43]. This graph structure information is crucial for understanding fault propagation paths and assessing the system-level impact of maintenance interventions, but traditional methods often fail to effectively utilize it [44].

In recent years, Graph Neural Networks (GNNs) have emerged as a powerful deep learning tool specifically designed for processing and learning graph-structured data, offering new avenues for addressing the aforementioned challenges. Graph-structured data is ubiquitous in the real world, such as social networks, molecular structures, and knowledge graphs [45-47]. The core mechanism of GNNs involves learning representations of nodes or edges in a graph through message passing between nodes or aggregating information from neighboring nodes. By stacking multiple GNN layers, models can capture both local and global structural information and attribute features of nodes within a graph, achieving significant success in various graph analysis tasks such as node classification, link prediction, and graph classification. GNNs have been widely applied in fields such as recommendation systems, signal processing, and traffic prediction. Jiang et al. summarized the rapidly growing research findings on using different graph neural networks (such as graph convolutional networks and graph attention networks) in various traffic prediction problems, including road traffic flow and speed prediction, passenger flow prediction in urban rail transit systems, and demand prediction in ride-hailing platforms [48].

Therefore, this study introduces GNN to explicitly model the interdependencies between devices in a manufacturing system. We abstract the system as a graph structure, where devices are nodes and the relationships between them are edges. By applying GNN, the model can learn a context-aware state representation for each device, which not only encodes the health status of the device itself but also incorporates the influence of the system environment in which it operates. This representation, rich in structured dependency information, provides a more comprehensive understanding of the system state for subsequent reinforcement learning strategy optimization, enabling the learning of more robust and globally aware maintenance decision strategies, and ultimately improving the long-term operational efficiency of the entire simulation manufacturing system.

2.4. Reinforcement learning for maintenance decision making

As discussed earlier, the precise state transition probabilities and cost functions of complex systems are often difficult to obtain. In such cases, Reinforcement Learning (RL) offers a powerful, model-free solution. RL enables agents to learn an optimal policy $\pi(a|s)$ through trial and error by directly interacting with the environment, selecting the optimal maintenance action $a$ given system state $s$ to minimize long-term cost [49]. In recent years, RL, particularly deep reinforcement learning (DRL), has been applied to maintenance optimization [50,51]. For example, Attestog et al. proposed a method combining a self-supervised anomaly detector based on local outlier factors (LOF) with a deep Q-network (DQN) supervised reinforcement learner [52]. This method can detect multiple faults in permanent magnet synchronous motors during dynamic operation without using historical labeled fault training data. Li et al. constructed a novel fault diagnosis model based on capsule neural networks (Cap-net) and proposed a novel online domain-adaptive learning method based on deep reinforcement learning to enhance the adaptability of the fault diagnosis model [53]. However, many existing DRL methods may face issues such as low sample efficiency and training instability when handling high-dimensional, strongly time-correlated sensor data specific to industrial scenarios, or they may still rely on traditional encoders such as PCA or standard VAE for state representation, which may fail to fully capture the intrinsic features of complex (e.g., non-Gaussian or multimodal) faults. Furthermore, most methods fail to explicitly consider the impact of inter-device dependencies on optimal maintenance strategies during modeling, which may lead to locally optimal rather than globally optimal decisions.

To overcome the shortcomings of traditional DRL methods in terms of stability and efficiency, researchers have proposed a series of advanced strategy optimization algorithms. Proximal Policy Optimization (PPO) is a widely used and robust strategy gradient algorithm in the field of reinforcement learning [54]. Liu et al. proposed a deep reinforcement learning method based on Multi-Agent Approximate Policy Optimization (MAPPO) to provide drone-assisted mobile edge computing (MEC) services [55]. The numerical results of this study demonstrate that the proposed scheme can effectively reduce energy consumption and outperform the baseline in terms of performance, convergence speed, and resource utilization. Compared with traditional policy gradient methods, PPO introduces a pruned alternative objective function or an adaptive KL divergence penalty term to limit the magnitude of each policy update, preventing excessive update steps from disrupting the learning process [56]. This mechanism enables PPO to achieve learning stability while typically offering simpler implementation and better sample efficiency than other methods, leading to state-of-the-art results in decision-making tasks with complex continuous or discrete action spaces, such as robot control and vehicle scheduling.

Given the good balance between performance, stability, ease of implementation, and sample efficiency of PPO, as well as its widespread success in complex decision-making tasks, this study selects PPO as the core reinforcement learning algorithm to learn the optimal maintenance strategy based on the system state representation proposed by us, which integrates temporal analysis (FNO), state representation (DAE), and graph structure information (GNN).

## 3. Preliminaries and Problem Statement

This section introduces the basic concepts, symbolic notation, and formal problem definitions used throughout this paper[1].

**Definition 1** ( Equipment State Space $S$ ). collects all possible conditions or health states of the monitored device or system. The specific state $s_t \in S$ at the decision moment $t$ contains information extracted from potentially high-dimensional, multi-source time series data, reflecting the current condition of the device.

**Definition 2** (Action Space $A$ ). represents a set of limited maintenance interventions available at each decision moment $t$. For example, $a_t \in A$ = {'take no action', 'minor maintenance', 'major maintenance', 'replacement'}. Selecting an action $a_t$ will affect the state transition of the system and generate corresponding costs.

**Definition 3** ( Decision Epochs $T$ ). represents discrete points in time $t = 0, 1, 2, \cdots$ at which the system status is evaluated and maintenance actions are selected. The time intervals between cycles can correspond to fixed time units or operational cycles.

**Definition 4** (State Transition Probability $P$ ). defines system dynamics, i.e., the probability $P(s_{t+1} | s_t, a_t)$ that the system will transition to state $s_{t+1} \in S$ in the next decision cycle, given the current state $s_t \in S$ and selected action $a_t \in A$. In real-world scenarios, especially considering complex degradation patterns influenced by operational history and maintenance, these probabilities are typically unknown and must be learned or estimated from data.

**Definition 5** (Reward Function $R$ ). defines the immediate reward $R(s_t, a_t, s_{t+1})$ or $R(s_t, a_t)$ associated with taking an action $a_t$ in state $s_t$ and transitioning to state $s_{t+1}$. This function typically includes the cost of maintaining the action $a_t$ itself, as well as operational costs associated with state $s_t$ or $s_{t+1}$, such as downtime costs or penalties associated with failure events. The optimization objective is related

---
[1] The definitions of MDPs are derived from Reference [4], [5], and [10].

to minimizing cumulative costs.

**Definition 6** (Maintenance Policy $\pi$). represents a strategy or decision rule that maps states to actions. A deterministic strategy is a function $\pi: S \to A$ that specifies the action $a_t = \pi(s_t)$ to take in the given state $s_t$. A stochastic strategy $\pi(a|s)$ specifies the probability of taking an action $a$ in a given state $s$.

**Definition 7** (Discount Factor $\gamma$). is used to discount future rewards relative to immediate rewards. It reflects a preference for near-term results or accounts for uncertainty in long-term predictions.

**Problem definition:** Given a time series data stream representing the operating status of manufacturing equipment and sensor readings $X = \{x_0, x_1, \ldots, x_t, \ldots\}$, as well as historical maintenance actions $A = \{a_0, a_1, \ldots\}$ and their results (such as failures and downtime), the goal is to determine an optimal maintenance strategy $\pi^*: S \to A$ that minimizes the expected total discount cost over an infinite time period. Mathematically, the goal is to find the $\pi^*$ shown as belows:

$$V^{\pi^*}(s) = \max_\pi E_\pi \left[ \sum_{t=0}^{\infty} \gamma^t R(s_t, a_t) | s_0 = s \right] \quad \forall s \in S \tag{1}$$

Among them, $s_t$ is derived from history $(X_t, A_{t-1})$, represents the state in the decision-making cycle $t$, and $E_\pi[\bullet]$ is expected to be calculated based on the state transition $P(s_{t+1}|s_t, a_t)$ guided by the strategy $\pi$.

## 4. Methodology

This study proposes a comprehensive MDP model for optimizing the maintenance strategy of intelligent manufacturing equipment. The framework effectively integrates several advanced AI techniques, including Fourier Neural Operator (FNO), Denoising Autoencoder (DAE), Proximal Policy Optimization (PPO), and Graph Neural Network (GNN). Through the in-depth integration and innovative application of the above methods, the constructed model aims to significantly improve the efficiency, accuracy and robustness of predictive maintenance decision-making, thus providing more reliable and efficient operation and maintenance support for intelligent manufacturing systems.

4.1 Overview of the MDP framework

The central goal of equipment maintenance is to develop a set of optimal decisions in an uncertain environment to minimize long-term operating costs (including failure repair costs, downtime losses, etc.) while maximizing equipment availability. To this end, we rigorously formalize the problem as a Markov Decision Process. an MDP consists of the quintuple shown in **Table 1**.

**Table 1** The components of the MDP

| Definition | Meaning |
|---|---|
| $S$ | State space |
| $A$ | Action space |
| $P(s'|s,a)$ | State transfer probability |
| $R(s,a,s')$ | Reward function |
| $\gamma$ | Discount factor |

The intelligent body senses the state of the device in state $s \in S$, performs a maintenance action $a \in A$, and subsequently the system transfers to a new state $s'$ with probability $P(s'|s,a)$ and receives an immediate reward $R(s,a,s')$. The goal of the intelligence is to learn an optimal policy $\pi^*: S \to A$ that maximizes the cumulative expected discounted reward from any initial state.

Our proposed FNO-DAE-GNN-PPO framework is built on MDP and consists of three functional modules. The overall architecture is shown in **Figure.1**, which is designed to address a specific key challenge in the process of optimizing equipment maintenance.

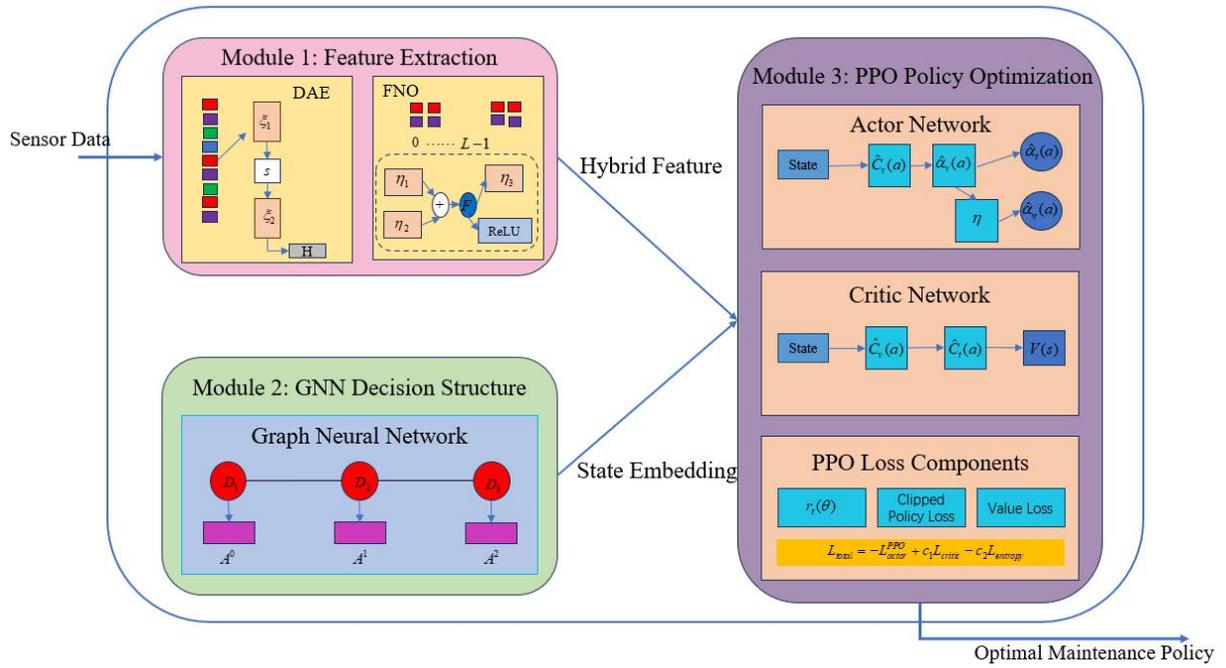

**Figure.1** The architecture of FNO-DAE-GNN-PPO MDP

**Module 1: Time-Series Processing and Frequency Domain Mapping**. This module aims to cope with the complex characteristics of raw time-series data generated by industrial equipment sensors, such as high frequency, non-stationarity, high dimensionality and strong noise interference. In order to realize the in-depth analysis and information extraction of this kind of data, this paper proposes a collaborative processing framework, which integrates the spectral analysis based on the Z-transform principle, the data characterization optimization driven by Denoising Autoencoder (DAE), and the in-depth time-series pattern learning supported by Fourier Neural Operator (FNO). Learning. The framework firstly denoises and compresses the original input by DAE to obtain a purer and more concentrated feature representation, which provides a stable basis for subsequent pattern recognition. Subsequently, the system extracts key features along two parallel paths: on the one hand, the Discrete Fourier Transform (DFT) based on the Z-transform principle is applied directly to the original or normalized data to efficiently capture the periodic structure and frequency domain energy distribution; on the other hand, the FNO models the data in the latent variable space extracted by the DAE, so as to learn the complex nonlinear dynamics and long-range dependency relationships. This multi-level and multi-perspective fusion strategy significantly improves the comprehensive modeling capability of equipment operation behavior patterns through refined nonlinear mapping and cross-domain feature transformation. Ultimately, the module provides robust and semantically rich feature support for downstream graph neural network (GNN) condition assessment and reinforcement learning-based proximal policy optimization (PPO) maintenance decisions.

**Module 2: Decision Structure - GNN**. In order to effectively model the complex topology and potential fault propagation paths among devices in the decision-making process, Graph Neural Network (GNN) is introduced in this module. Based on the low-dimensional state representation output from DAE, GNN combines the connection relationship between devices and learns to generate a state embedding with structure-aware capability. The embedding not only retains the state information of individual devices, but also encodes their contextual dependencies in the overall network, thus providing richer and more structured input features for the policy network.

**Module 3: Policy Optimization - PPO.** As the core module of the reinforcement learning framework, this part uses the Proximal Policy Optimization (PPO) algorithm to jointly train a stochastic policy network (Actor) and a value estimation network (Critic) with the graph embedding states generated by the GNN as input. network (Actor) and a value estimation network (Critic). The Actor network outputs the probability distribution of each maintenance action in the current state, which is used to guide the action selection, while the Critic network evaluates the value of the current state, which provides a reference signal for the policy to update.PPO and effectively controls the policy update amplitude through the introduction of a trimmed objective function, which improves the stability of the training process and monotonicity of the convergence.

The following subsections elaborate on the design philosophy, mathematical principles, implementation details and their role in the overall framework of each module.

## 4.2. Timing Processing and Frequency Domain Mapping Module

Time-series data generated by devices is a key basis for assessing their health status and predicting potential failures. However, such data typically exhibits complex characteristics such as high dimensionality, high noise, nonlinearity, and nonsmoothness, posing significant challenges for modeling and analysis. Traditional time-domain analysis methods are often limited in capturing long-term dependencies and subtle anomaly patterns. In contrast, frequency domain analysis can reveal the periodic structure and energy distribution of signals, providing a complementary and insightful perspective for troubleshooting.

### 4.2.1. Spectral feature based on Z-Transform principle

The purpose of using the Z-Transform is to capture the underlying frequency-domain properties directly from the original sensor time series. The Z-Transform serves as the cornerstone of discrete signal processing, converting the time-series signal $x[n]$ into a function $X(z)$ in the complex Z-plane. In particular, when the Z-Transform is evaluated on the unit circle ($z = e^{j\omega k}$), it is equivalent to the Discrete Time Fourier

Transform (DTFT), which is discretely sampled as the Discrete Fourier Transform (DFT). The DFT is able to reveal the energy distribution of a finite-length sequence at different discrete frequency points. Therefore, this branch quantifies the intensity of the DFT coefficients of each channel of the original signal at selected frequency components by calculating their amplitudes, which is crucial for identifying fault indication messages with specific frequency signatures.

For each independent sensor channel $s$, the original time series $X_s = [x_s(t_0), x_s(t_1), \ldots, x_s(t_{L-1})]$ are subjected to the following processing steps.

(1) Real Sequence Fast Fourier Transform (RFFT): Considering that the input $X_s$ is a sequence of real numbers, in order to improve the computational efficiency and to utilize the conjugate symmetry of the spectrum, a one-dimensional real Fast Fourier Transform (RFFT) is used to compute its one-sided spectrum. This operation produces $[L/2]+1$ complex Fourier coefficients, denoted as $\hat{X}_s[k]$, for an input of length $L$, where $k = 0, 1, \ldots, [L/2]$. Each coefficient $\hat{X}_s[k]$ contains amplitude and phase information corresponding to the frequency $f_k = k \cdot (f_{sample}/L)$ (if the sampling rate $f_{sample}$ is known). The standard DFT calculation formula is as follows:

$$\hat{X}_s[k] = \sum_{n=0}^{L-1} x_s(t_n) e^{-j\frac{2\pi}{L}kn} \tag{2}$$

(2) Calculation of spectral amplitude: For each complex Fourier coefficient $\hat{X}_s[k]$, we calculate its mode to obtain the amplitude of the corresponding frequency component $M_s[k] = |\hat{X}_s[k]| = \sqrt{\operatorname{Re}(\hat{X}_s[k])^2 + \operatorname{Im}(\hat{X}_s[k])^2}$. This magnitude directly reflects the concentration of energy in the signal at that frequency point.

(3) Feature vector construction and dimension unification: From the computed sequence of magnitudes $[M_s[0], M_s[1], \ldots, M_s[L/2]]$, we select the first $N_{Z_{feat}}$ amplitudes as the final spectral signature of this sensor channel. If the number of available magnitudes is less than $N_{Z_{feat}}$, it is padded with zeros at the end of the sequence to ensure dimensional consistency; if it is more than that, only the first $N_{Z_{feat}}$ is retained. This portion of the data is typically the low to mid-frequency portion, as many of the eigenfrequencies of mechanical failures lie in this region. This ensures that each channel outputs a fixed dimension of features.

(4) Numerical Stability Guarantee: All non-numerical or infinite values in the extracted eigenvalues are replaced with 0.0 to ensure the stability of the subsequent computations. The above steps are performed for all $N_{sensors}$ sensor channels, and the $N_{Z_{feat}}$ dimensional spectral feature vectors obtained from each of them are spliced to form the final output feature $f_{z\_spectral}$ of this branch, whose total dimensionality is $N_{sensors} \times N_{Z_{feat}}$.

4.2.2. Deep temporal pattern based on Fourier Neural Operator (FNO)

This branch aims to mine and model more complex nonlinear global dynamical patterns as well as long-range dependencies in time series, features that are often difficult to capture by traditional linear frequency domain analysis or simple time domain statistical methods. To this end, we introduce the Fourier Neural Operator (FNO), a deep neural network architecture for function-space learning, whose core innovation is to efficiently learn the mapping relationship between input and output functions by parameterizing the integral kernel operator in the Fourier domain. The method effectively captures long-range dependencies by realizing point-by-point multiplication in the frequency domain and thus performing global convolution in the function space. Compared with traditional neural networks, FNO does not rely on a specific form of data discretization and has the potential to handle sequences of different lengths and sampling rates, thus demonstrating stronger generalization ability and modeling robustness when facing diverse time-series data structures in industrial equipment. **Figure.2** shows the flowchart of FNO.

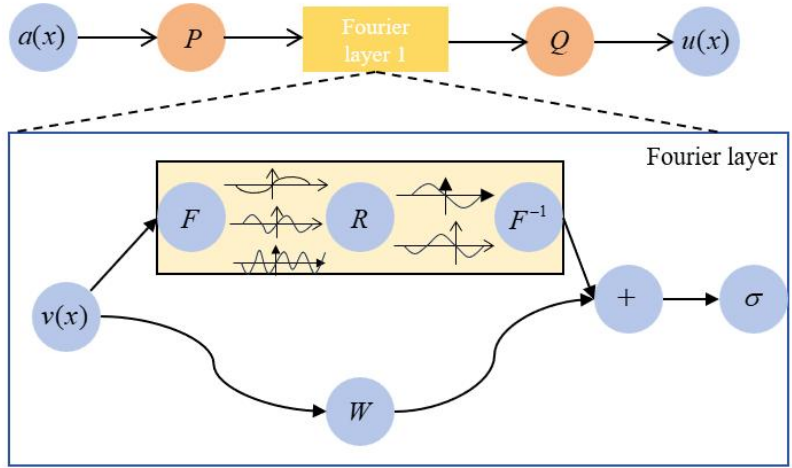

**Figure.2** The flowchart of FNO

Raw sensor data are usually characterized by high dimensionality, strong noise, and potential covariance, which may lead to inefficient learning, difficult model convergence, or overfitting to noisy details if directly input into the Fourier Neural Operator (FNO). In order to improve the learning performance of the FNO branch and enhance its ability to model the intrinsic dynamic structure of the data, this study introduces the Denoising Autoencoder (DAE) in its front-end for preprocessing and feature characterization.

The DAE uses an encoder, decoder architecture for unsupervised training with the optimization objective of minimizing the reconstruction error between $\hat{x}(t)$ reconstructed from the addition of a Gaussian noise input $x_{noisy}(t)$ and the original clean input $x(t)$, formalized as:

$$L_{DAE} = \left\| x(t) - D_{DAE}(E_{DAE}(x_{noisy}(t))) \right\|_2^2 \qquad (3)$$

This training mechanism motivates the encoder $E_{DAE}$ to learn noise-insensitive latent representations.

Specifically, the multichannel sensor data $x(t) \in R^{N_{sensors}}$ at each time step is mapped to a low-dimensional, information-dense latent vector $z(t) \in R^{d_{DAE}}$.

$$z(t) = E_{DAE}(x(t)), \text{ where } d_{DAE} < N_{sensors} \tag{4}$$

As a result, the sensor sequence $X = [x(t_0),\ldots,x(t_{L-1})]$ over the whole observation window is transformed into the latent vector sequence $Z_{DAE} = [z(t_0),\ldots,z(t_{L-1})]$. This latent representation not only has lower dimensionality and higher signal-to-noise ratio, but also can highlight the key information more prominently, which significantly improves the efficiency and robustness of FNO in modeling complex temporal structures.

Based on the latent feature input of DAE, FNO performs feature extraction. Where the input is the DAE latent vector sequence $Z_{DAE} \in R^{L \times d_{DAE}}$. The process is carried out in four parts.

(1) Channel Lifting: The sequence $V_0 \in R^{L \times d_w}$ is obtained by a fully connected layer applied to each time-step latent vector $z(t_i)$ in the sequence, which is linearly mapped from the $d_{DAE}$ dimension and lifted to the channel dimension $d_w$ required for the internal processing of FNO. The data is then typically reshaped into a tensor format to accommodate subsequent convolution-like operations.

(2) Optional Padding: We perform tail padding on the sequence $V_0$ in the time dimension, with the purpose of dealing with boundary effects that may arise from subsequent convolution operations, or to ensure that sequences of different lengths can have consistent internal representation lengths after processing in the FNO layer.

(3) Iterative processing of $FNO1D$ layers: Sequence $V_0$ passes through the same $FNO1D$ layers of $N_{FNO}$ sequentially. Each $FNO1D$ layer ($l = 0,\ldots,N_{FNO}-1$) performs the following core operation on its input $V_l$ (with dimensions $(B, d_w, L')$ and $L'$ is the length of the filled sequence)

(a) Spectral convolution paths: a. Fourier Transform ($F$): Apply RFFT to each channel of $V_l$ ($d_w$ in total) along the time dimension (length $L'$) to obtain its frequency domain representation $\hat{V}_l \in C^{B \times d_w \times ([L'/2]+1)}$. b. Frequency domain linear transformation (filtering and interaction): In the Fourier domain, only the first $k_{max}$ low frequency modes are selected. For these selected modes, a learnable complex weight tensor $R_l \in C^{d_w \times d_w \times k_{max}}$ is applied to linearly combine them. Specifically, for each output channel $j$ and each selected mode $m$ ($0 \leq m < k_{max}$):

$$\left(\hat{v}_l'\right)_j[k_m] = \sum_{p=1}^{d_w}(R_l)_{jp,m} \cdot (\hat{v}_l)_p[k_m] \tag{5}$$

This operation allows the model to learn how the frequency components of the different input channels are combined to form the output channel's frequency components, enabling cross-channel frequency domain information interaction and weighting. The unselected high frequency modes are usually set to zero. c. Inverse Fourier Transform ($F^{-1}$): The weighted combined and truncated frequency-domain representation $\hat{V}'_l$ (with the high-frequency portion zeroed out to match the full spectral length) is converted back to the time domain by the IRFFT to obtain the output of the spectral path, $V'_{spec,l} \in R^{B \times d_w \times L'}$.

(b) Localized linear paths: In parallel, a $1 \times 1$ one-dimensional convolution (with weights $W_l \in R^{d_w \times d_w \times l}$) is applied to the input $V_l$ of the $FNO1D$ layer. This is equivalent to applying a linear transformation to the $d_w$ channel features once at each time step, capturing local, cross-channel feature interactions, often as a residual connection or complementary path. The output $V'_{lin,l} \in R^{B \times d_w \times L'}$ is obtained.

(c) Path fusion and activation: The output of the spectrally convolved path $V'_{spec,l}$ is summed element-by-element with the output of the locally linear path $V'_{lin,l}$, and then the final output of the current $FNO1D$ layer $V_{l+1}$ is obtained by a nonlinear activation function:

$$V_{l+1} = GELU(V'_{spec,l} + V'_{lin,l}) \tag{6}$$

Sequence flattening and final projection: After processing all $N_{FNO}$ and $FNO1D$ layers, the final feature sequence $V_{N_{FNO}}$ (with dimensions $(B, d_w, L')$) is flattened into a long vector (with dimensions $B \times (d_w \cdot L')$). This vector is then projected through a fully connected layer and a ReLU activation function to a predefined FNO branch output dimension, forming the final feature vector $f_{FNO}$ for that branch.

To form a unified and comprehensive description of the device state, we effectively fuse the heterogeneous feature vectors extracted from the above two parallel branches. The first branch is the DFT-based spectral feature vector $f_{Z\_spectral}$ in Section 3.2.1, and the second branch is the FNO-based deep temporal pattern feature $f_{FNO}$ in Section 3.2.2. These two feature vectors are directly spliced in the feature dimension, and the method is able to retain the specific information extracted from each branch intact. The pseudocode for this module is shown as **Algorithm 1** below.

**Algorithm 1**: Time Processing and Frequency Domain Mapping Module

**Input**: Sensor data
**Output**: Hybrid features
1:      function TimeFrequencyMapping(S):
2:      /* Branch 1: Z-Transform spectral feature extraction */
3:          f_spectral = []
4:          **For** each channel c in S do
                fourier_coeffs = REFT(S[:, c, :])
                amplitudes = |fourier_coeffs|[: n_freq]

|       |                                                                 |
|-------|-----------------------------------------------------------------|
|       |          f.spectral.append(amplitudes)                          |
|       |     **End**                                                     |
| 5:    |     f_spectral = Concatenate(f_spectral)                        |
| 6:    |     /* Branch 2: DAE encoding and FNO processing */             |
| 7:    |     encoder = TrainedDAE_Encoder()                              |
| 8:    |     Z = [encoder(S[:, :, t])] for t in range(seq_length)        |
| 9:    |     v = FNO_Process(Z)                                          |
| 10:   |     f_fno = ProjectAndFlatten(v)                                |
| 11:   |     f_hybrid = Concatenate([f_spectral, f_fno])                 |
| 12:   |     return f_hybrid                                             |

Depending on the needs of subsequent tasks and considerations of model complexity, this raw hybrid feature vector can be optionally passed through a final fully-connected projection layer, often suffixed with a ReLU activation function, to generate the final hybrid feature module output $f_{hybrid}$. This output will be used as an input node feature to the downstream model (the graph neural network GNN in this study), providing it with rich information about the state of the device within the current observation window.

4.3 Decision architecture module based on Graph Neural Network (GNN)

After completing the aforementioned hybrid module to extract the individual feature vectors $f_{hybrid}$ of each device, this paper further introduces Graph Neural Network (GNN) to model and mine the potential correlations and interactions among groups of devices. In real industrial environments, equipment usually does not operate independently, and its operating state may be influenced by other equipment in the same production unit, shared operating environment, or upstream and downstream processes. Therefore, relying only on individual characteristics may not fully reflect the real operating state of a device, GNN learns the structured representation of each device in the overall system by constructing a graph structure with devices as nodes and their physical connections, functional couplings, or fault propagation paths as edges, and by using a message passing mechanism to propagate and integrate the information among the device neighborhoods. This mechanism is able to generate state embeddings with global context-awareness, effectively enhancing the model's expressive ability and generalization performance in multi-device collaborative maintenance decision-making.

To introduce the GNN, we first formally model the device system as a graph $G = (V, E, X)$, where each node $v_i \in V$ in the node set $V$ represents an individual physical device whose initial node feature vector $x_i$ is the output $f_{hybrid}$ generated by the aforementioned hybrid feature extraction module. The feature vectors of all devices together form the node feature matrix $X \in R^{N \times D_{hybridfinal}}$, where $N$ denotes the total number of devices and $D_{hybridfinal}$ is the dimension of the hybrid feature vector. The edge $e_{ij} \in E$ in the graph is used to represent the predefined association relationship between devices $v_i$ and $v_j$. In this study, this association is defined based on whether the devices belong to the same business unit or not: if two devices belong to the same group, a powerless and directionless edge is added between them. Based on this criterion, an adjacency matrix $A \in \{0,1\}^{N \times N}$ can be constructed. In order to enhance the stability and expressiveness of the subsequent graph convolutional network (GCN) during the training process, we symmetrically normalize the adjacency matrix $A$ to obtain the normalized adjacency matrix $\bar{A} = D^{-1/2}AD^{-1/2}$, where the $D$ is the $A$ degree matrix, defined as $D_{ii} = \sum_{j} A_{ij}$.

In this study, Graph Convolutional Network (GCN) is used as a concrete implementation of GNN to learn the enhanced representation of the device nodes (Wang). GCN updates the representation of the central node by iteratively aggregating the feature information of the neighboring nodes.

GCN Layer Propagation Rule: For a GCN model containing an $L_{GCN}$ layer, the node feature matrix $H^{(l+1)} \in R^{N \times d^{(l+1)}}$ of the $(l+1)$ th layer is computed from the feature matrix $H^{(l)} \in R^{N \times d^{(l)}}$ of the $l$ th layer in the following way (where $H^{(0)} = X$):

$$H^{(l+1)} = \sigma(\bar{A}H^{(l)}W^{(l)}) \quad (7)$$

Where $W^{(l)} \in R^{d^{(l)} \times d^{(l+1)}}$ is the $l$ th layer learnable weight (parameter) matrix and $\sigma(\bullet)$ is the nonlinear activation function.

(1) Initial node representation input ($H^{(0)}$): The computation of the GCN model starts at layer 0, and its input $H^{(0)}$ is the node feature matrix $X \in R^{N \times D_{hybridfinal}}$ outputted by the aforementioned "timing processing and frequency domain mapping module" for the $N$ devices in the system. The input $H^{(0)}$ is the node feature matrix $X \in R^{N \times D_{hybridfinal}}$, where $D_{hybridfinal}$ is the dimension of the hybrid feature vector $f_{hybrid}$ of each device after integrating the time domain, frequency domain and deep learning modes.

(2) Hidden Layers ($l = 0, \ldots, N_{GCN\_layers} - 2$): The model contains $N_{GCN\_layers} - 1$ hidden GCN layers. Each hidden layer $l$ receives the node representation $H^{(l)}$ of the previous layer and computes its output $H^{(l+1)}$ as follows. The first step is feature transformation and aggregation. First, the input features $H^{(l)}$ are linearly transformed by a learnable weight matrix $W^{(l)} \in R^{d^{(l)} \times d^{(l+1)}}$. According to the common practice of GCNs, especially when the adjacency matrix is normalized, this linear transformation does not introduce bias terms (Fang). Subsequently, the transformed features are weighted and aggregated using the normalized adjacency matrix $\bar{A}$, which is essentially a process where each node collects information about its neighboring nodes. The second step is nonlinear activation: the aggregated features are then subjected to a ReLU nonlinear activation function $\sigma(\bullet)$ in order to introduce nonlinear expressiveness, enabling the model to learn more complex inter-node relationships. Therefore, the output $H^{(l+1)}$ of the $l+1$ st hidden layer is calculated as:

$$H^{(l+1)} = \mathrm{Re}LU(\bar{A}H^{(l)}W^{(l)}) \quad (8)$$

The input dimension $d^{(0)}$ of the first hidden layer ($l=0$) is $D_{hybrid\_final}$, and its output dimension $d^{(1)}$, as well as the output dimensions of all subsequent hidden layers (if $N_{GCN\_layers} - 1 > 1$) are $d_{hidden}$. This means that the node feature dimension is mapped from $D_{hybridfinal}$ to $d_{hidden}$ in the first hidden layer and maintains this dimension in the subsequent hidden layer processing up to the output layer.

(3) Output Layer: The last GCN layer is responsible for linearly projecting the high-level abstract representation of a node to the final target embedding dimension $d_{output}$. It is computed as:

$$E_{GNN} = \overline{A} H^{(N_{GCN\_layers}-1)} W^{(N_{GCN\_layers}-1)} \tag{9}$$

where $W^{(N_{GCN\_layers}-1)} \in R^{d_{input\_to\_output\_layer} \times d_{output}}$ is the weight matrix of the output layer. A key design choice here is that this output layer does not employ a nonlinear activation function. This treatment aims to preserve the full range of values of the embedding vectors and their potential linear differentiability, allowing them to more adequately reflect the subtle differences between the states of the nodes. In reinforcement learning, the embedding vectors will be used as state inputs to the intelligences, so maintaining their original expressiveness is a common and effective practice to help avoid the information compression and nonlinear distortion that activation functions may bring. The pseudocode for this module is shown as **Algorithm 2** below.

| **Algorithm 2**: GNN-Based Decision Architecture Module |
|---|
| **Input**: Hybrid features; device groups |
| **Output**: Node embeddings |
| 1:    Function GNNDecisionArchitecture(f_hybrid, G): |
| 2:        A = ZerosMatrix(n_devices, n_devices) |
| 3:        **For** each group g in G do |
|              **For** each pair (i,j) in g where i≠j do |
|                  A[i,j] = 1 |
|              End |
|           End |
| 4:        D_inv_sqrt = DiagonalMatrix([1/√Sum(A[i,:]) for i in range(n_devices)]) |
| 5:        A_norm = D_inv_sqrt · A · D_inv_sqrt |
| 6:        H = f_hybrid // Initial node features |
| 7:        **For** l = 1 to L-1 do |
|              H = ReLU(A_norm · H · W^(l)) |
|           End |
| 8:        E = A_norm · H · W^(out) |
| 9:        Return E |

By stacking graph convolutional layers, the final embedded representation of each node is able to recursively aggregate information about its features in up to $N_{GCN\_layers}$ hopping neighborhoods in the device association graph. As the number of layers increases, the node will have the ability to perceive a larger range of graph structures, thus capturing wide-area inter-device dependencies and system-level effects. The generated embedding vectors $e_i$ not only concisely characterize the complex features of the device itself (inherited from the hybrid representation $f_{hybrid}$), but also incorporate its structural location and neighborhood context information in the whole device topology, providing a more expressive and globally-aware state representation for the subsequent PPO intelligences.

### 4.4. Strategy optimization module based on PPO

After obtaining a finely constructed device state representation from the GNN module, the core task of this study turns to learning optimal maintenance policies based on this state representation. To this end, we choose the Proximal Policy Optimization (PPO) algorithm as the core of the reinforcement learning framework, which is a widely used and well-performed policy gradient method in current deep reinforcement learning and is widely recognized for its stability, implementation simplicity, and good real-world performance in the policy update process.

#### 4.4.1. Strategy gradient and the Actor-Critic framework

PPO belongs to the Policy Gradient (PG) method. PG methods directly parameterize the policy $\pi_\theta(a|s)$, which means the probability of choosing an action $a$ in state $s$, where $\theta$ is a parameter of the policy network and maximize a certain performance metric, $J(\theta)$, by a gradient ascent. PPO is usually implemented in an Actor-Critic framework.

Actor policy network ($\pi_\theta(a|s)$) learns and executes a policy. The input is the state $s$ (in this framework the state embedding $h_G$ for the GNN output) and the output is the probability distribution over the action space for discrete actions or the parameters of continuous actions.

Critic value Network ($V_\phi(s)$ or $Q_\phi(s,a)$) is used to the learning and evaluation of the value function aims to quantify the potential payoffs of states or state-action pairs to aid the optimization process of the policy network. Specifically, the introduction of the value function helps to reduce the variance of the policy gradient estimation, thus improving the stability and efficiency of training. In this study, we construct the state-value function $V_\phi(s)$ and the action-value function $Q_\phi(s,a)$, which are used to estimate the expected payoffs that can be obtained by following the current policy in state $s$ and the expected payoffs that can be obtained by continuing to follow the current policy after performing action $a$ in state $s$, respectively, and both of them are defined by the parameter set $\phi$.

#### 4.4.2. Core mechanisms of PPOs

Considering the high cost of actual interactions in industrial manufacturing systems, we construct offline datasets based on historical records. Specifically, we normalize and encode (DAE) the sensor data from multi-device multi-temporal sequences, extract the raw and latent features corresponding to the time windows, and generate device-level temporal representations via the Z-FNO hybrid feature extractor. At each time step $t$, we utilize the current representations of all devices to construct the graph neural network inputs and compute their graph embedding representations, where the node embedding of the target device is taken as the state $s_t$ at that time step. Subsequently, we combine the historical action record $a_t$, the cost $c_t$ and the next moment state $s_{t+1}$ to construct the hexadecimal experience:

$$(s_t, a_t, \log \pi_{\theta_{old}}(a_t | s_t), r_t, s_{t+1}, d_t) \tag{10}$$

where the reward is defined as $r_t = -c_t$ and the termination flag $d_t$ is determined based on the device failure state.

In order to control the magnitude of policy updates and maintain training stability, we use the PPO's tailoring objective function. In each round of training iteration s, we first re-evaluate the policy probabilities on the old samples using the current policy and calculate the policy ratio:

$$r_t = \exp\left(\log \pi_\theta(a_t | s_t) - \log \pi_{\theta_{old}}(a_t | s_t)\right) \tag{11}$$

Then we introduce the time-differential target value and define the dominance function accordingly:

$$\hat{Q}_t = r_t + \gamma V(s_{t+1}) \tag{12}$$

$$A_t = \hat{Q}(t) - V(s_t) \tag{13}$$

The dominance values are further normalized to reduce estimation noise.

Next, the PPO strategy loss function is defined as:

$$L_{actor}^{PPO} = E_t\left[\min\left(r_t A_t, clip(r_t, 1-\varepsilon, 1+\varepsilon) A_t\right)\right] \tag{14}$$

where $\varepsilon$ is the trimming factor used to constrain the policy update not to deviate from the old policy. The loss in the Critic branch is the mean-square error loss of the state-value function:

$$L_{critic} = E_t\left[\left(V(s_t) - \hat{Q}_t\right)^2\right] \tag{15}$$

Also we add entropy regularization terms to encourage strategy diversity:

$$L_{entropy} = E_t[\mathrm{H}(\pi_\theta(\cdot | s_t))] \tag{16}$$

This results in a total loss function:

$$L_{total} = -L_{actor}^{PPO} + c_1 L_{critic} - c_2 L_{entropy} \tag{17}$$

where $c_1$ and $c_2$ control the balance of weights between value function learning and strategy exploration, respectively. The pseudocode for this module is shown as **Algorithm 3** below.

**Algorithm 3**: PPO-Based Strategy Optimization Module

**Input**: Experience data; parameters

**Output**: Optimized policy

```
1:      Function PPOOptimization(D, ε, γ):
2:          π_θ = InitializeActor(state_dim, action_dim)
3:          V_φ = InitializeCritic(state_dim)
4:          For epoch = 1 to n_epochs do
                batches = CreateMiniBatches(Shuffle(D))
            End
6:          For batch in batches do
                s_t, a_t, r_t, s_{t+1}, done_t = batch
                log_π_old = LogProb(π_θ(s_t), a_t)
                A_t = ComputeAdvantages(r_t, V_φ(s_t), V_φ(s_{t+1}), done_t, γ)
                For k = 1 to K do
                    ratio = exp(LogProb(π_θ(s_t), a_t) - log_π_old)
                    L_clip = min(ratio·A_t, clip(ratio, 1-ε, 1+ε)·A_t)
                    L = -L_clip + c₁·(V_φ(s_t) - (A_t + V_φ(s_t)))² - c₂·Entropy(π_θ(s_t))
                    UpdateParameters(π_θ, V_φ, L)
                End
            End
7:          Return π_θ
```

Policy updates are in the form of mini-batches, where the entire experience pool is randomly disrupted and trained in batches in each iteration, and $K$ policy updates are repeatedly executed for each mini-batch to fully utilize the data and achieve progressive policy updates. At the beginning of each iteration, the current policy log probabilities of all actions are recalculated and used as inputs for estimating the old policy distribution in the current update. This mechanism effectively avoids strategy collapse caused by sample bias and improves strategy learning stability. In our experiments, we select $K=4$, use up to 1024 empirical samples for training in each round, and continuously optimize the policy network parameters in 20 rounds of main loops, and finally obtain a stable converged policy function $\pi^*(a|s)$.

## 5. Experimental results

### 5.1. Data and presetting

#### 5.1.1. Data

Given that large-scale, long-term operational data containing comprehensive fault evolution information is difficult to obtain directly in real industrial environments due to its commercial sensitivity, high acquisition costs, and low frequency of fault events, the use of simulation-generated representative datasets has become a key research paradigm for algorithm validation and strategy optimisation in the field of predictive maintenance. Therefore, this study constructed a synthetic time series dataset to simulate the operational characteristics of equipment throughout its entire lifecycle [57,58]. This study employed a synthetic time series dataset designed for developing and evaluating predictive and prescriptive maintenance strategies. The dataset simulated the operational behaviour of 50 independent devices throughout their respective lifecycles, with each device containing 2,000 records of discrete time steps, totaling 100,000 data points. Each record represents the state of a specific device at a specific point in time, identified by a unique equipment ID and an incrementing time step. Additionally, devices are grouped into five equipment groups, providing a foundation for introducing graph structure information to analyse potential correlations among devices within the same group. This structured synthetic data facilitates the training and testing of complex maintenance decision models in a controlled environment. Each record in the dataset contains multi-dimensional features that comprehensively describe the operational state and maintenance history of the device. State features are primarily composed of five simulated sensor readings: wear level, vibration, temperature, pressure, and error rate. These continuous-value variables aim to capture dynamic changes in device health status and potential pre-failure indicators. Action data includes four normalised maintenance options: do nothing, minor maintenance, major maintenance, and replace. Corresponding to these actions, the cost incurres column quantifies the direct economic cost of each maintenance intervention, where do nothing costs $0, minor maintenance costs $100, major maintenance costs $500, and replace costs $2,000.The last column of the dataset marks whether the device has failed at the current time step in binary form. The overall failure rate simulated in this dataset is 2%, which is primarily set to provide sufficient failure samples for the predictive maintenance model to support effective training and evaluation. This simulated dataset aims to provide a representative and controllable experimental platform. Its generation characteristics include: all devices have a uniform trajectory length, ensuring data consistency for sequence modelling methods; in terms of action distribution, 'do nothing' accounts for approximately 75% of the actions, simulating the scenario where devices are mostly in normal operation or observation mode in actual operations; sensor data is designed to reflect the process of devices transitioning from normal operation to gradual degradation and ultimately leading to failure in some cases. Although the simulated data may not fully replicate all the complexity and noise of the real world, it provides a clear and reproducible foundation for systematically studying maintenance strategies, validating model performance, and exploring the impact of different parameters on decision-making behaviour. These characteristics make the dataset highly suitable for training and evaluating the predictive maintenance decision models based on deep learning and reinforcement learning proposed in this study.

5.1.2 Evaluation index

To comprehensively evaluate the performance of the proposed method in the task of optimising the maintenance of intelligent manufacturing equipment, this paper selects three core evaluation indicators: minimum long-term average cost, strategy quality, and computation time. The minimum long-term average cost measures the comprehensive cost of maintenance and downtime per unit time during the execution of the strategy in long-term operation, reflecting the economic efficiency of the strategy in actual industrial applications. The evaluation of the optimal strategy focuses on the rationality and stability of action selection under different states, typically measured indirectly through the performance of the strategy after convergence. Computation time records the code execution time of each method during training and inference, reflecting the engineering feasibility and deployment efficiency of the model.

5.1.3 Hyperparameter setting and selection methods

To ensure that all submodules operate efficiently and collaboratively in the maintenance strategy learning task, this paper systematically sets and optimises key hyperparameters in the deep reinforcement learning framework, covering the clipping threshold for PPO strategy updates, the latent representation dimension of DAE, the number of information aggregation layers in GNN, and the number of frequency domain truncation modes in FNO. For these parameters, this paper adopts a hybrid strategy combining grid search and experience-based settings for optimisation, and verifies their performance impact through experiments.

(1) PPO clipping parameter ($\varepsilon$): In the PPO algorithm, the clipping parameter $\varepsilon$ is a core hyperparameter that defines the trust region for the change amplitude between the old and new strategies. By limiting the proportion of strategy updates, this parameter effectively prevents strategy collapse caused by excessive single update steps, thereby enhancing the stability of the training process. This paper initially adopts the default value (0.2) proposed by Gu et al. [59], followed by a grid search within the range $[0.1, 0.25]$ with a step size of 0.05. The variability of the strategy convergence curves and the final average cost under different settings are examined, ultimately confirming that 0.2 performs optimally under most task conditions.

(2) DAE latent dimensions: DAE is used to extract low-dimensional latent feature representations from raw high-dimensional sensor data. The 'latent dimensions' parameter directly determines the dimension size of the compressed feature space. A higher latent dimension can retain more information from the original data, but may also increase the computational complexity and overfitting risk of subsequent models. In contrast, a lower dimension aims to learn more compact and robust features. In this paper, the potential dimension is set within the range of 16, 32, 64, and 128. Models are trained under each configuration, and the average long-term cost and computation time are evaluated. Ultimately, 32 dimensions are selected as the trade-off point between performance and efficiency.

(3) Number of GNN layers: GNN learns the representation of each node in the graph by aggregating information from neighbouring nodes. The number of GNN layers determines the 'hop count' or range of information propagation in the graph, i.e., how many layers of neighbouring information can be incorporated into the final representation of each node. However, in the specific implementation of the system, a basic number of layers may first be adopted based on the network scale and task complexity, and then adjusted through subsequent experiments to achieve the best context understanding ability while avoiding over-smoothing caused by too many layers. Considering that three layers are widely adopted in the literature [60,61] and effectively avoid over-smoothing issues, this paper conducts comparative experiments on 2-layer, 3-layer, and 4-layer structures and finds that the three-layer network achieves the best balance between policy convergence and computational complexity.

(4) FNO frequency domain truncated mode number: When processing time series data, FNO captures different frequency patterns by performing convolution operations in the frequency domain. The 'number of frequency domain truncated modes' refers to the number of low-frequency modes retained after the Fourier transform. Retaining a larger number of modes allows the model to capture finer high-frequency changes in the data, making it suitable for complex dynamic systems. In practical applications, a smaller number of modes may be selected based on the spectral characteristics and noise level of the data to focus on the main dynamic trends and improve the model's generalisation ability. This parameter is also configurable to optimise it according to data characteristics. Through FFT analysis, we observed that the majority of effective frequencies are concentrated within the first 60–70 frequency points. After conducting comparative experiments within the range of 32, 64, and 128, we selected 64 as the default value to balance expressive capability and computational overhead.

During parameter tuning, grid search is used to explore discrete combinations of key hyperparameters. The evaluation criteria include training stability, convergence iteration count, average cost performance, and inference latency. For some parameters (such as PPO clipping threshold and GNN layer number), since there is already a large body of literature and reinforcement learning practices that provide effective references, this paper combines experience to select initial values and verifies their reasonableness within a small range. For example, the clipping coefficient 0.2 is a common default value in many reinforcement learning studies and has been proven to effectively balance strategy exploration and stability in continuous action spaces. Similarly, the 3-layer design of GNN is also consistent with the common scenario of medium-complexity industrial equipment structures.

To further validate the sensitivity and robustness of the selected hyperparameters to model performance, this paper conducts a systematic hyperparameter sensitivity analysis experiment in Section 5.6, evaluating their impact on policy optimality and training dynamics from multiple dimensions.

5.2 Baseline model comparison

To ensure that the maintenance strategy is executable and engineering-oriented, we have discretised the degradation process of the equipment into seven state levels, denoted as Wear_0 to Wear_6, which correspond to different degrees of wear from 'healthy' to 'serious failure.' The specific classification is based on the degradation indicators in the monitoring data and historical maintenance records, which are determined through cluster analysis and expert experience, as shown in **Table.1**.

Table.1 Classification of equipment status and corresponding explanations

| **Equipment status** | **Explanation** |
| --- | --- |
| Wear_0 (New/Healthy State) | The equipment is operating stably with no obvious abnormalities. |
| Wear_1 (Minor Wear) | Initial signs of minor degradation have appeared. |
| Wear_2 (Moderate Wear) | Equipment performance has slightly declined. |
| Wear_3 (Moderate Deterioration) | Production stability is already affected |
| Wear_4 (Severe Deterioration) | The risk of failure has significantly increased. |
| Wear_5 (Near Failure) | The equipment is nearly at the point of functional failure. |
| Wear_6 (Complete Failure) | The equipment has already failed or lost all functionality. |

This grading system takes into account the actual operating status of the equipment and helps to clearly express the model strategy. In the strategy comparison, the action selection of different methods at different Wear levels reflects their understanding of the degradation law and maintenance optimisation level.

To comprehensively evaluate the performance of the FNO-DAE-GNN-PPO MDP framework proposed in this paper, we selected three representative benchmark models for comparison. These models represent different modelling ideas and computational paradigms, which are described in detail below.

**MDP + Q-learning**: This method models the equipment wear process as a discrete Markov decision process and uses Q-learning for strategy learning. This method is based entirely on tabular reinforcement learning and does not use any state abstraction techniques, so it has weak generalisation ability but can serve as a basic comparison for traditional reinforcement learning.

**DQN-PCA MDP**: This method introduces deep Q-networks (DQN) into the traditional MDP framework for policy fitting, while combining principal component analysis (PCA) to reduce the dimensionality of the state space, thereby enhancing the ability to handle high-dimensional states. Compared to Q-learning, DQN has stronger non-linear fitting capabilities, while PCA helps extract key features and improve model convergence.

**Transformer-VAE MDP**: This model utilises the Transformer architecture for time series modelling and variational autoencoders (VAE) for latent representation learning of the degradation process. This method aims to capture complex temporal features but lacks explicit environment modelling and reinforcement learning optimisation mechanisms, relying more on the abstract capabilities of unsupervised representations, resulting in limited policy performance.

**Table.2** shows the experimental results of the three baseline models and the FNO-DAE-GNN-PPO MDP model proposed in this paper.

From the perspective of minimising long-term average cost, as shown in Table X, the FNO-DAE-GNN-PPO MDP proposed in this paper achieves the best performance in terms of long-term average maintenance cost, at only \$91.95 (per time step), significantly outperforming all baseline methods. The Q-learning and DQN-PCA methods have the same minimum long-term average cost, both at \$105.70 (per time step), while the Transformer-VAE MDP achieves a minimum long-term average cost as high as \$3000.00 per time step, indicating that its strategy is nearly ineffective in controlling the degradation process. This result demonstrates that the proposed method, through the integration of joint modelling and reinforcement learning optimisation, can more accurately predict the future state of equipment, reasonably schedule maintenance plans, and thereby achieve superior cost control.

From the perspective of strategy structure, the strategies learned by Q-learning and DQN-PCA MDP are relatively conservative, following traditional maintenance logic: choosing not to maintain during mild wear, and gradually adopting minor repairs, major repairs, and replacement measures as wear increases. In contrast, the DQN-PCA model delays replacement decisions at higher wear levels and tends to continue using maintenance strategies, which may be attributed to state information loss caused by the dimensionality reduction process. The strategy learned by the Transformer-VAE MDP selects 'no action' at all wear levels, representing a clear failure strategy. This indicates that the model fails to effectively map degraded states to reasonable maintenance actions, potentially influenced by modeling biases in latent variables and the absence of reinforcement signals. In contrast, the strategy learned by FNO-DAE-GNN-PPO is more refined and gradual, demonstrating precise control over the degree of equipment degradation while ensuring reliability and avoiding premature or excessive maintenance. FNO provides high-precision spatio-temporal prediction capabilities,

DAE achieves robust state feature extraction, GNN enhances the perception of device structural relationships, and PPO ensures the stability and performance improvement of strategy updates. In terms of computation time, Transformer-VAE MDP is the shortest at 5.68 seconds, followed by DQN-PCA and Q-learning at 6.03 seconds and 7.02 seconds, respectively.

The proposed method has a relatively long computation time of 92.19 seconds due to its complex model structure and the strategy optimisation process involving multiple simulations and iterations. However, in real-world industrial scenarios, maintenance strategies are typically planned on a daily or weekly basis, making this computational cost still acceptable.

Overall, while traditional methods offer some interpretability and computational efficiency, they lack sufficient generalisation and modelling capabilities when dealing with complex wear evolution patterns. Deep representation methods such as Transformer-VAE suffer from unstable strategy performance. The proposed FNO-DAE-GNN-PPO framework effectively integrates multiple advanced models, forming a closed-loop system in state modelling, structural representation, and strategy optimisation. It achieves significant advantages in maintenance cost, strategy rationality, and system reliability, demonstrating broad industrial application prospects.

Table.2 Relevant experimental results of the four models

|  | Minimum Long-term average cost | Optimal strategy | Calculation time |
|---|---|---|---|
| MDP+Q-learning | $105.70 (per time step) | Wear_0: do nothing<br>Wear_1: minor maintenance<br>Wear_2: major maintenance<br>Wear_3: major maintenance<br>Wear_4: replace<br>Wear_5: replace<br>Wear_6: replace | 7.02s |
| DQN-PCA MDP | $105.70 (per time step) | Wear_0: do nothing<br>Wear_1: minor maintenance<br>Wear_2: major maintenance<br>Wear_3: major maintenance<br>Wear_4: major maintenance<br>Wear_5: replace<br>Wear_6: replace | 6.03s |
| Transformer-VAE MDP | $3000.00 (per time step) | Wear_0: do nothing<br>Wear_1: do nothing<br>Wear_2: do nothing<br>Wear_3: do nothing<br>Wear_4: do nothing<br>Wear_5: do nothing<br>Wear_6: do nothing | 5.68s |
| FNO-DAE-GNN-PPO MDP | $91.95 (per time step) | Wear_0: do nothing<br>Wear_1: minor maintenance<br>Wear_2: minor maintenance | 92.19s |

|  |
|---|
| Wear_3: major maintenance |
| Wear_4: major maintenance |
| Wear_5: major maintenance |
| Wear_6: replace |

5.3 Analysis of Model Computational Complexity

To evaluate the scalability and computational efficiency of the maintenance optimization framework proposed in this paper in engineering applications, this paper conducts a item-by-item analysis of the computational complexity of its core modules, including FNO, DAE, GNN, and PPO. Moreover, a comprehensive comparison is made with the Typical Baseline Method (DQN).

In the state modeling part, the FNO module uses the Fast Fourier Transform (FFT) for sequence modeling in the frequency domain. Its single-layer computational complexity mainly includes two parts: $o(N \log N)$ of the FFT operation, where $N$ is the length of the time series, and the complexity of the frequency-domain filtering calculation is approximately $o(N \cdot K)$, where $K$ is the number of truncated frequency-domain modes. Therefore, for L-layer FNO, the total complexity is $o(L(N \log N + N \cdot K))$. During the state compression stage, the DAE module consists of a symmetrical fully connected encoder-decoder. Let the input feature dimension be $d_{in}$ and the latent representation dimension be $d_z$. Then the complexity of each forward propagation is $o(d_{in} \cdot d_z)$, which is relatively low overall and can effectively compress the state dimension and accelerate subsequent processing.

In the graph structure modeling stage, the GNN module uses a three-layer graph convolutional network, and each layer performs adjacency aggregation operations on nodes. Let the number of nodes in the graph be $n$, the average number of neighbors per node be $k$, and the feature dimension of the nodes be $d$. Then the complexity of the convolution of each layer of the graph is $o(n \cdot k \cdot d)$, and the three-layer network in this study is $o(3nkd)$. In the strategy generation section, the PPO module involves the forward and backward propagation of the strategy and value function network. Let the batch size be $B$ and the number of parameters in the policy network be $P$, then the complexity of each round of policy update is approximately $o(B \cdot P)$.

Overall, the total computational complexities of each training iteration of the method proposed in this paper are as follows.

FNO: $o(L(N \log N + N \cdot K))$

DAE: $o(d_{in} \cdot d_z)$

GNN: $o(3nkd)$

PPO: $o(B \cdot P)$

In contrast, the DQN method usually employs perceptrons or convolutional neural networks for state modeling, lacking the capabilities of frequency domain modeling and graph structure modeling. Its single-round training complexity is mainly composed of forward propagation and backpropagation, generally being $o(B \cdot P')$, where $P'$ is the number of network parameters, which is usually less than the sum of the two networks in the PPO structure. However, due to its limited state modeling ability, the quality of the strategy is difficult to guarantee in high-dimensional and complex states.

Therefore, although the complexity of the method proposed in this paper is slightly higher in the training stage, it can effectively capture the global frequency domain structure of state evolution, the topological relationship between nodes, and the long-term policy value function, and has better economy and policy stability in actual maintenance tasks. Meanwhile, in the deployment stage, through means such as model pruning, frequency-domain truncation optimization, and GNN sparse adjacency representation, the reasoning efficiency can be further improved to meet the real-time requirements of industry.

5.4 Convergence analysis

During the training process of the PPO algorithm, the convergence of the strategy is one of the key indicators for measuring the performance and training effect of the algorithm. This study evaluates the convergence of the PPO algorithm by tracking the changing trend of KL divergence during the strategy update process. The KL divergence measures the degree of difference between the old and new strategies in each iteration. A smaller KL divergence value usually indicates a smaller update amplitude of the strategy and a more stable strategy.

**Figure.3** shows the average approximate KL divergence of the policy after each iteration (including 4 K-epochs sub-updates) in 20 PPO training iterations. The following trends can be clearly observed in the figure.

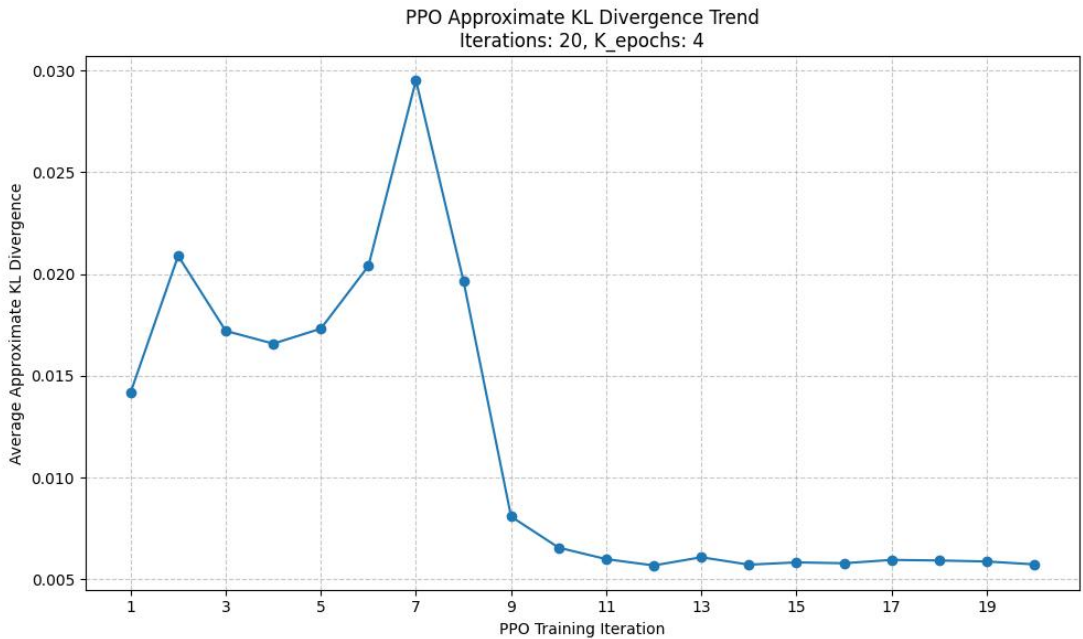

**Figure.3** The KL divergence trend of PPO

Firstly, the period from the first to the eighth iteration is the initial exploration and significant adjustment phase. During the initial training stage, the KL divergence value is relatively high and exhibits significant fluctuations. For example, at the second iteration, the KL divergence reached approximately 0.021, and at the seventh iteration, it reached the peak of the entire training process, approaching 0.030. This indicates that during the initial training phase, the PPO agent is actively exploring the state space and action space, and the policy network undergoes significant adjustments and optimisations, with strong learning signals. This phenomenon aligns with the learning characteristics of reinforcement learning algorithms in the initial stage, where large strategy updates are used to rapidly search for potential high-quality strategy regions.

moreover, the period from the 9th to the 20th iteration represents the rapid convergence and strategy stabilisation phase. Starting from the 8th iteration, the KL divergence value shows a sharp downward trend. After the 9th iteration, the KL divergence rapidly decreased to below 0.010 and remained at a relatively stable low level (approximately 0.005 to 0.006) during subsequent iterations (iterations 10–20). This significant decrease in the KL divergence value and its eventual stabilisation at a low level strongly indicates that the PPO strategy has converged to a stable state. The agent has found a relatively optimised strategy, and subsequent iterations primarily involve fine-tuning around this strategy, with no significant changes to the strategy itself.

Throughout the training process, the trend of KL divergence changes in the PPO strategy clearly demonstrates the complete process from initial exploration and adjustment to stable convergence in the later stages. The sustained low KL divergence values in the later stages of training indicate that the trained PPO agent has successfully converged to a stable strategy, laying a solid foundation for subsequent decision analysis and performance evaluation.

5.5 Ablation experiment

In order to validate the independent contributions and overall synergistic effects of each key module in the FNO-DAE-GNN-PPO model proposed in this paper, we further designed four sets of ablation experiments, which respectively removed the FNO, DAE, GNN, and PPO modules, and compared the changes in long-term average cost, optimal strategy, and computation time among the models. The specific results are shown in **Table.3**.

From the perspective of long-term average cost, the complete model achieves the lowest value of $91.95 (per time step), significantly outperforming variants with missing modules. The model lacking the GNN module exhibits the most significant cost increase ($105.25 per time step), approaching the traditional DQN-PCA baseline model ($105.70 per time step), indicating that graph structural information plays a crucial role in modelling the interdependencies and degradation propagation characteristics between device components. The model without the FNO module has a cost of $94.15 (per time step), slightly higher than the complete model, suggesting that FNO's advantages in spatio-temporal modelling capabilities can further enhance the foresight and economic efficiency of the strategy. Removing the DAE results in only a slight increase in cost ($93.95 per time step),but computational time decreased, indicating that its primary role lies in feature noise reduction and stable state representation. Removing PPO, while maintaining the cost at an optimal level ($92.50 per time step), resulted in increased computational time and strategy volatility, suggesting that PPO remains indispensable for ensuring the smoothness of strategy updates and improving convergence quality.

From a strategy structure perspective, the action selections of various ablation models in the Wear_0 to Wear_4 stages are similar to those of the complete model, indicating that state recognition in this interval is relatively reliable across multiple modules. However, in the Wear_5 and Wear_6 stages, there are significant differences in the strategies of different models. For example, the model without GNN directly selects 'replace' at Wear_5, indicating its inability to capture the state delay effects between components, leading to the premature adoption of high-cost maintenance strategies. Meanwhile, the model without PPO, although its strategy is similar to the complete model, may experience strategy drift or instability during multiple iterations due to the lack of constraints on the strategy update process.

In summary, the four submodules in this method perform their respective functions and collaborate in modelling. FNO provides dynamic modelling capabilities, DAE improves state expression stability, GNN enhances structure awareness, and PPO ensures strategy optimisation efficiency and stability. The complete model achieves the best balance between maintenance costs, strategy rationality, and computational overhead through inter-module collaboration mechanisms, verifying the effectiveness and necessity of its design.

Table.3 The Ablation experiment of FNO-DAE-GNN-PPO MDP model

|  | Minimum Long-term average cost | Optimal strategy | Calculation time |
|---|---|---|---|
| Model without FNO | $94.15 (per time step) | Wear_0: do nothing<br>Wear_1: minor maintenance<br>Wear_2: minor maintenance<br>Wear_3: major maintenance<br>Wear_4: major maintenance<br>Wear_5: major maintenance<br>Wear_6: replace | 149.91s |
| Model without DAE | $93.95 (per time step) | Wear_0: do nothing<br>Wear_1: minor maintenance<br>Wear_2: minor maintenance<br>Wear_3: major maintenance<br>Wear_4: major maintenance<br>Wear_5: major maintenance<br>Wear_6: replace | 82.27s |
| Model without GNN | $105.25 (per time step) | Wear_0: do nothing<br>Wear_1: minor maintenance<br>Wear_2: minor maintenance<br>Wear_3: major maintenance<br>Wear_4: major maintenance<br>Wear_5: replace<br>Wear_6: replace | 74.14s |
| Model without PPO | $92.50 (per time step) | Wear_0: do nothing<br>Wear_1: minor maintenance<br>Wear_2: minor maintenance<br>Wear_3: major maintenance<br>Wear_4: major maintenance | 125.67s |

|   |   |
|---|---|
|   | Wear_5: major maintenance |
|   | Wear_6: replace |

5.6 Hyperparameter sensitivity analysis

We conduct sensitivity analyses on the hyperparameters corresponding to DAE, FNO, GNN, and PPO.

**Figure.4** represents the hyperparameter sensitivity analysis of DAE. The sensitivity analysis of the DAE module indicates that its performance is primarily influenced by the noise level of the input data, the selected latent dimension, and the learning rate. The charts show that lower noise levels (0.3) and appropriately selected latent dimensions (16) lead to better convergence and faster convergence speed, while the optimal learning rate (0.001) is crucial for balancing convergence speed and final loss. Other DAE parameters such as batch size and epochs exhibit stronger robustness. Smaller batch sizes and more epochs generally improve loss, but the effect diminishes with parameter adjustments, and overall sensitivity is lower than that of noise, latent dimension, or learning rate. Regarding other model components, analysis of sequence length indicates that longer sequences generally improve overall scores and feature quality, but runtime increases significantly after reaching an optimal point (approximately length 8).

In **Figure.5**, for FNO's parameters, both FNO modes and FNO width exhibit an inverted U-shaped relationship with performance metrics such as overall score and feature quality, indicating an optimal solution (both set to 36). Beyond this range, performance may decline or runtime may increase.

In **Figure.6**, regarding the GNN, the sensitivity analysis focused on the number of information aggregation layers, a key factor determining the range of information propagation and the model's ability to understand context while avoiding over-smoothing. Comparative experiments are conducted using 2-layer, 3-layer, and 4-layer GNN structures. The findings indicate that a 3-layer GNN achieves the optimal balance between policy convergence and computational complexity, aligning with common practice in the literature for handling medium-complexity industrial equipment structures and effectively preventing over-smoothing issues.

For the PPO algorithm in **Figure.7**, the sensitivity of the PPO clipping parameter ($\epsilon$), which controls the extent of policy updates and training stability, is systematically evaluated. A grid search is performed within the range of 0.1 to 0.3 (step size 0.05), starting from a default value of 0.2. The analysis of strategy convergence and final average cost across these settings confirm that the clipping parameter value of 0.2 performed optimally under most task conditions, effectively balancing strategy exploration and stability, consistent with established reinforcement learning practices.

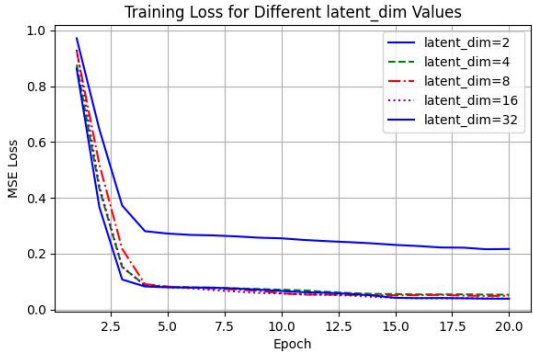
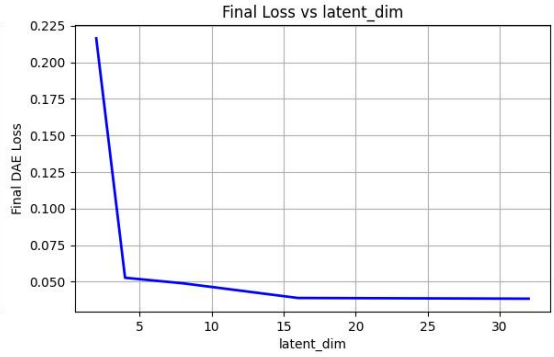
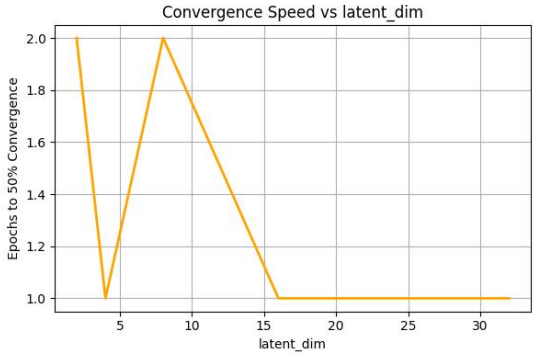
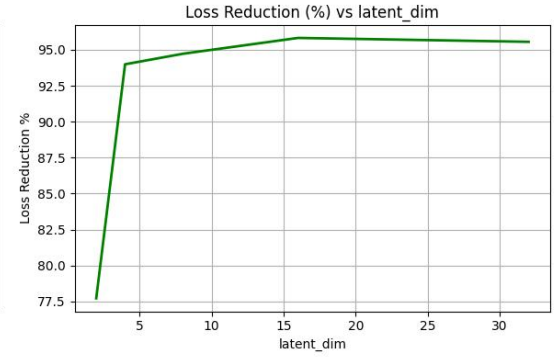
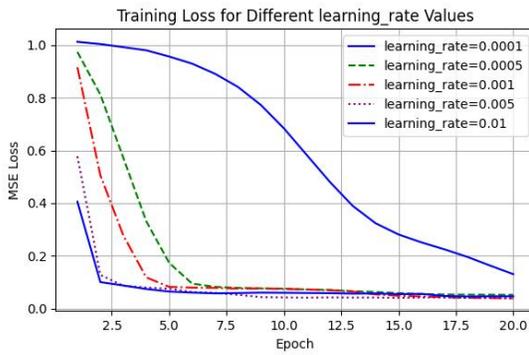
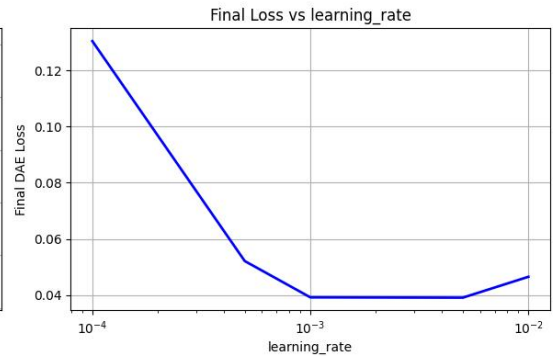
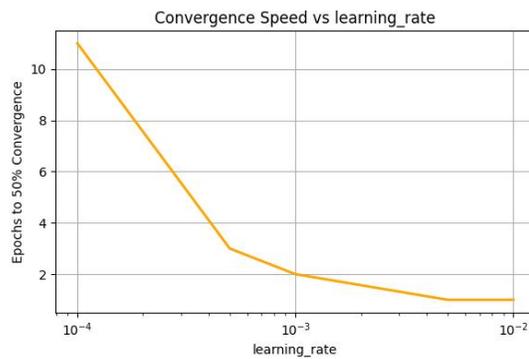
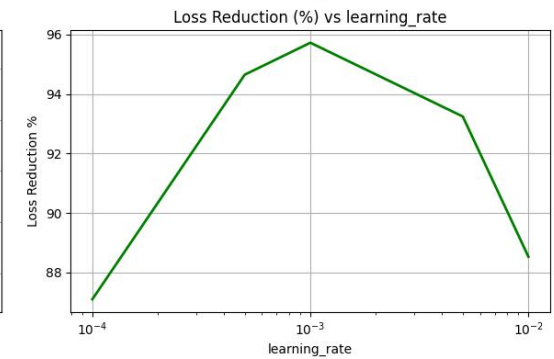

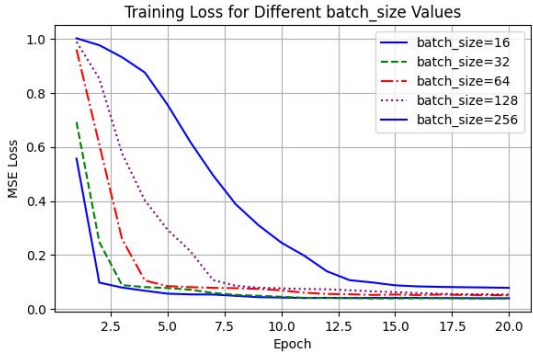
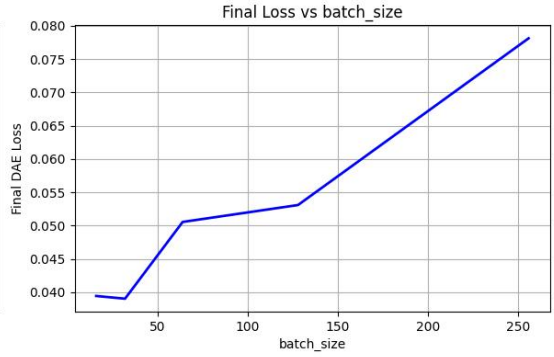
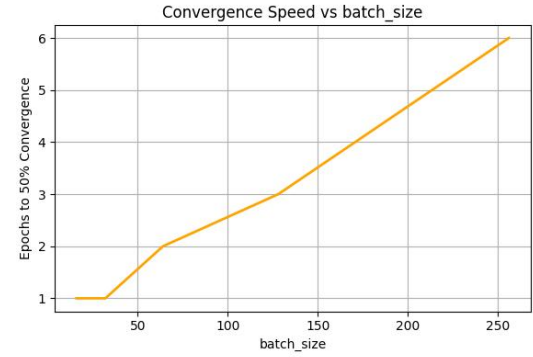
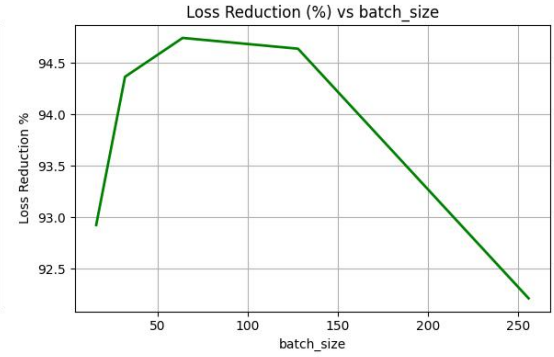
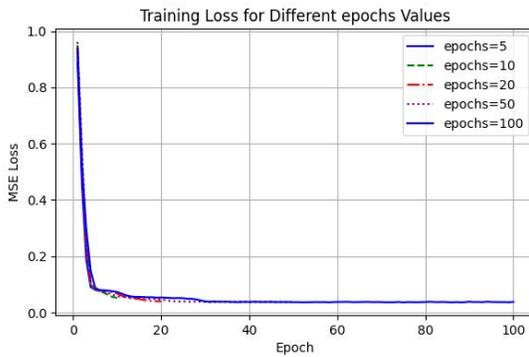
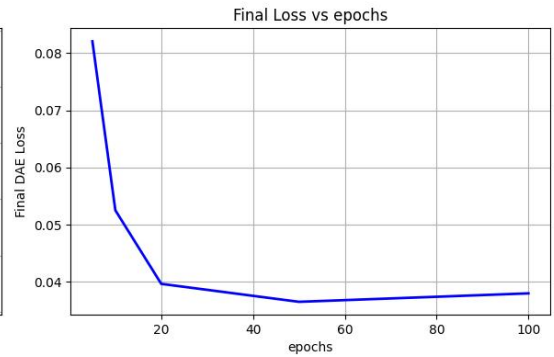
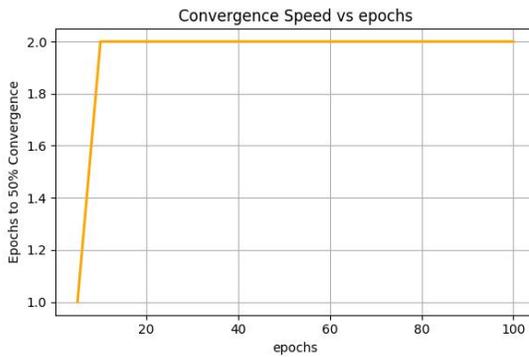
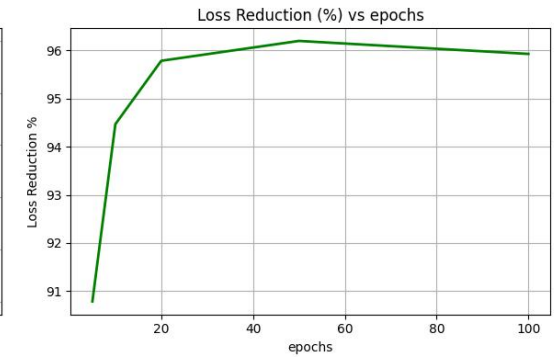

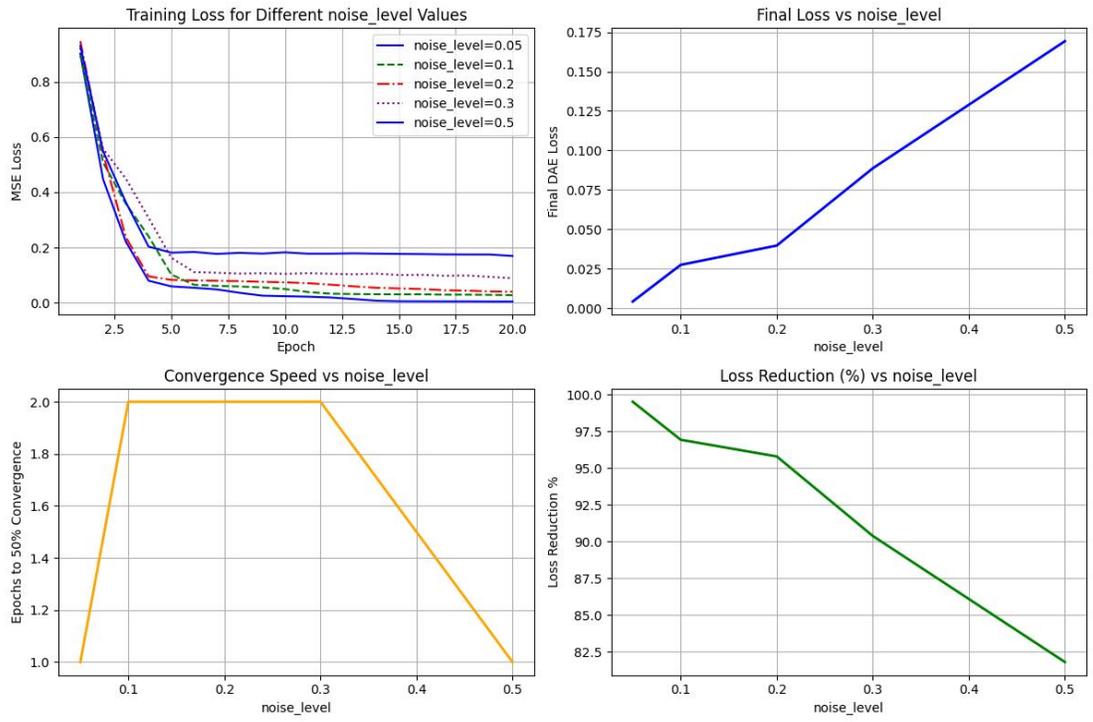

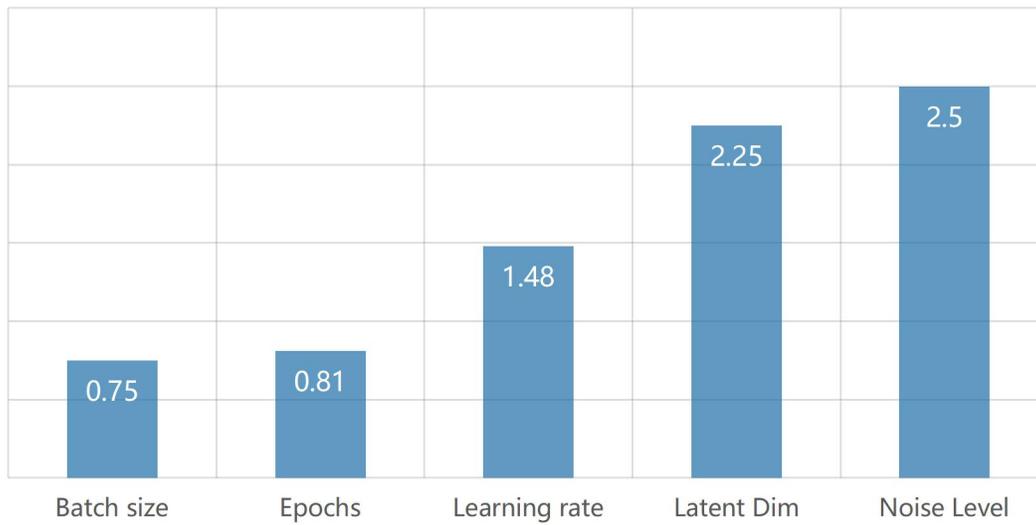

**Figure.4** The hyperparameter sensitivity analysis of DAE

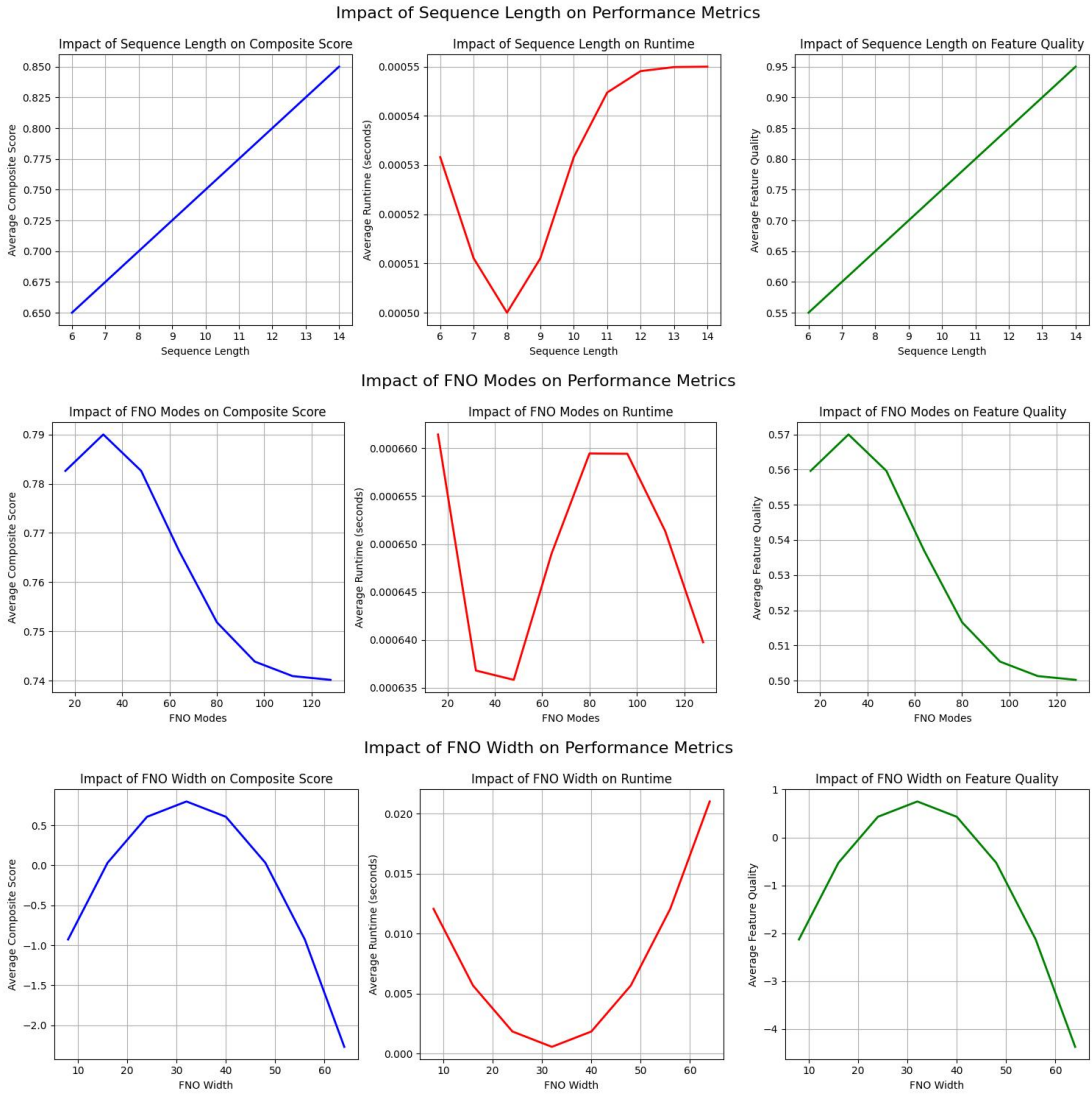

**Figure.5** The hyperparameter sensitivity analysis of FNO

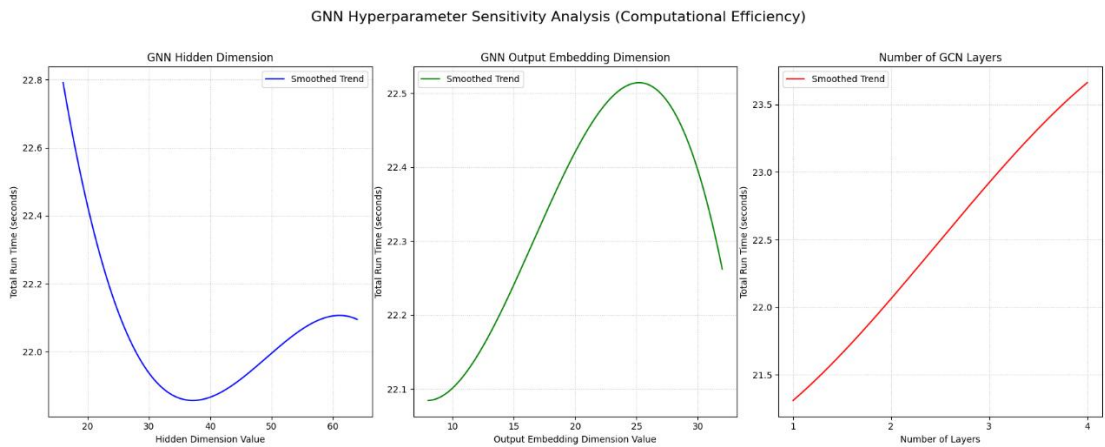

**Figure.6** The hyperparameter sensitivity analysis of GNN

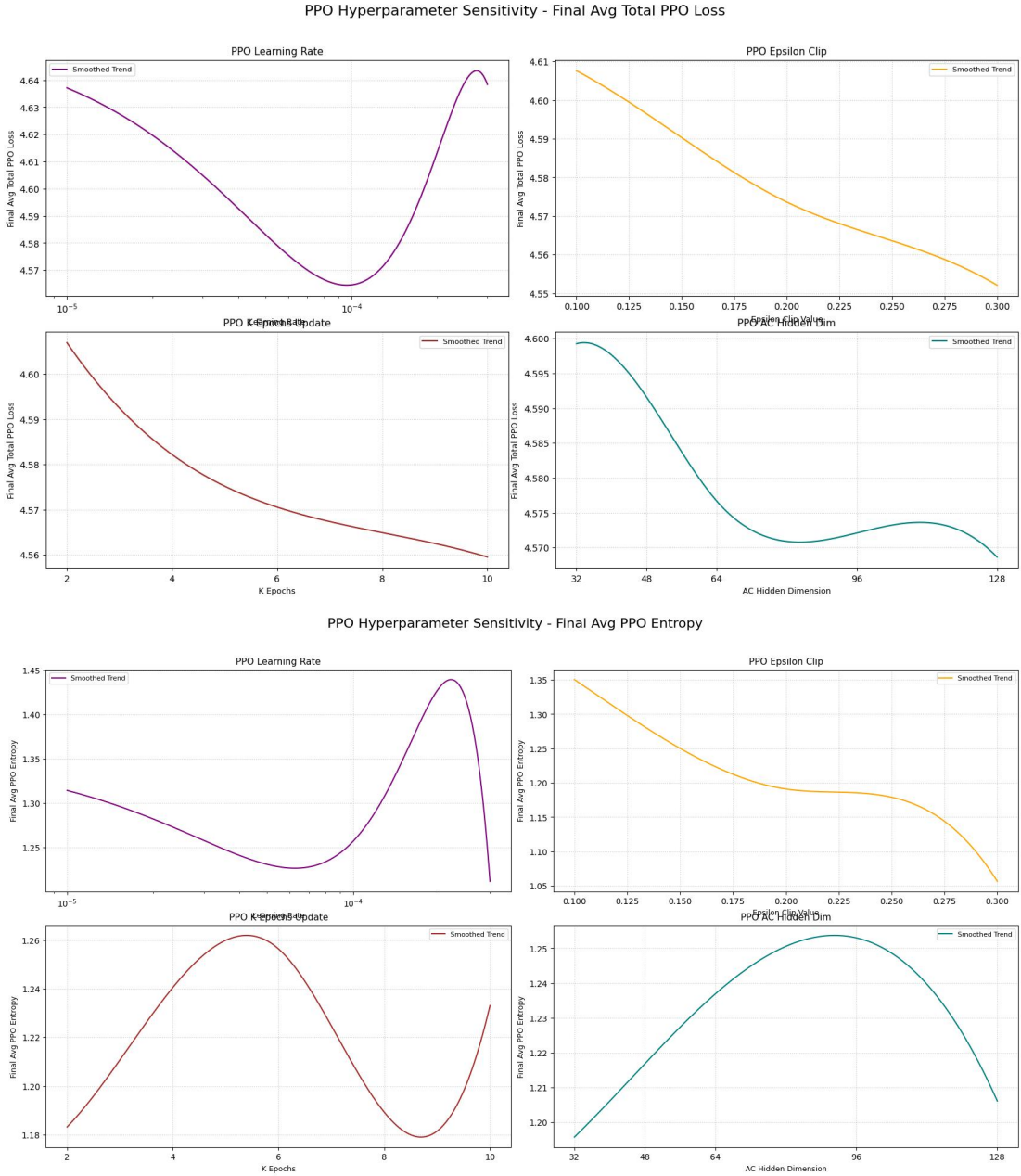

**Figure.7** The hyperparameter sensitivity analysis of PPO

5.7 Industrial data validation

After completing the model validation based on simulation data, our study further adopts the "Case Western Reserve University (CWRU) 12k Drive End Bearing Fault Data" in a real industrial environment for performance evaluation. The effectiveness of the use of FNO and DAE in real industrial scenarios is verified for the characteristics of bearing faults in the frequency domain that usually manifest as local energy concentration or harmonic resonance in specific frequency bands. The Fourier Neural Operator (FNO) is used as the core frequency domain feature extractor, which effectively captures and models such non-local and non-linear frequency domain dynamics through parametric learning in Fourier space. The features extracted by the FNO are then fed into the Noise-Absorbing Auto-Encoder (DAE) for in-depth processing, which is designed to learn highly robust low-dimensional latent representations.The core functionality of the DAE module is to significantly The core function of the DAE module is to significantly suppress the extraneous fluctuations and sensor noise prevalent in real industrial data, while maximising the preservation of the most discriminative structural features embedded in the fault signals.

**Table.5** Experiments on CWRU 12k Drive End Bearing Fault Data

| DAE epoch | Denoising MSE |
|---|---|
| Epoch 10 | 0.006565 |
| Epoch 20 | 0.003844 |
| Epoch 30 | 0.002153 |
| Epoch 40 | 0.001042 |
| Epoch 50 | 0.000398 |

**Table.5** represents the increasing denoising capability on noisy FNO features. It has performed perfectly in processing noisy FNO features with the average error decreasing from the $10^{-3}$ order of magnitude to the $10^{-4}$ order of magnitude. The resulting DAE encoder outputs highly robust low-dimensional latent representations that can be used for subsequent GNN and PPO solving of optimal policies.

## 6. Discussion and future work

This section delves into a comprehensive analysis of the experimental findings presented in Section 5, elaborates on the potential industrial applications and implications of the proposed FNO-DAE-GNN-PPO MDP framework, acknowledges the limitations inherent in the current study, and finally, outlines promising avenues for future research.

6.1 Analysis of findings

The experimental results robustly demonstrate the efficacy of the proposed integrated framework for optimizing predictive maintenance strategies.

Our FNO-DAE-GNN-PPO model achieves a significantly lower long-term average maintenance cost compared to established baseline methods.  This underscores the framework's capability in learning more cost-effective maintenance policies. The learned optimal strategy exhibits a more granular and adaptive approach to maintenance decisions across different equipment wear states.  Unlike baseline models that often resorted to overly conservative or ineffective strategies, our model demonstrates a balanced approach, recommending minor or major maintenance appropriately before resorting to replacement.  This suggests a better understanding of the degradation process and the trade-offs involved.

The ablation studies are crucial in validating the individual contributions of the FNO, DAE, GNN, and PPO modules. The absence of GNN leads to the most substantial increase in maintenance costs, highlighting the critical role of modeling inter-device dependencies for system-level optimization. Removing FNO also results in increased costs, confirming its value in capturing complex temporal dynamics for better state prediction. While removing DAE had a smaller impact on cost, its role in noise reduction and stable feature representation likely contributes to the overall robustness and efficiency observed. The PPO algorithm proves essential for stable and efficient policy optimization, as its removal, while maintaining a relatively good cost, increases computation time and potential policy volatility.

The convergence analysis, indicated by the KL divergence trend during PPO training, showes that the policy effectively stabilized after an initial phase of exploration, typically within 8-9 major iterations. This confirms the learning stability provided by the PPO algorithm within our framework. While the proposed framework demonstrated superior performance, it also incurres a higher computational time during training compared to the baseline methods. However, as maintenance strategies are often planned on a daily or weekly basis, this computational overhead is generally acceptable in practical industrial scenarios, especially given the potential long-term cost savings. The complexity analysis also details the contributions of each module to the overall computational load.

6.2 Potential in industrial applications

The FNO-DAE-GNN-PPO MDP framework holds considerable promise for transforming maintenance practices in intelligent manufacturing environments.

The framework can serve as a powerful decision support tool for maintenance planners, providing data-driven recommendations for when and what type of maintenance to perform on specific equipment within a complex production system. This moves beyond traditional scheduled or reactive maintenance towards truly predictive and prescriptive approaches. By optimizing maintenance interventions, the system can significantly reduce overall operational costs. This includes minimizing direct maintenance expenses and, more importantly, reducing indirect costs associated with unexpected downtime and production losses. The achieved average cost of \$91.95 per time step in simulations indicates strong potential for economic benefits.

Proactive and optimized maintenance ensures that equipment operates closer to its optimal performance levels for longer periods, thereby increasing overall equipment effectiveness and system reliability. A key strength is the GNN component's ability to consider inter-device dependencies. In many industrial settings, the health and maintenance of one machine can impact connected machines or the entire production line. Our framework can help identify and manage these systemic risks, leading to more globally optimal maintenance schedules. The integration of FNO and DAE allows the system to effectively process high-dimensional, noisy, and time-correlated sensor data, which is characteristic of modern industrial IoT environments. This makes the framework adaptable to various types of equipment and sensor inputs.

6.3 Limitations of the current method

Despite the encouraging results achieved, the current research still has certain limitations.

Firstly, the current verification relies on synthetic datasets. Although designed to be representative, industrial data in the real world typically exhibits greater complexity, including more diverse failure modes, non-stationarity, missing data, and unmodeled external influences. A thorough study on the transformation and performance of real data is required.

Secondly, the method in our paper integrates complex deep learning models, such as FNO and GNN, which may lead to a long delay in model inference, making it difficult to meet real-time decision-making requirements. To cope with this problem, techniques such as model pruning, knowledge distillation or quantisation can be used to reduce model complexity and improve inference speed. This method involves high-dimensional feature processing and multi-layer graph-structured networks, with high memory requirements, which puts greater pressure on the limited resources of edge devices.

Thirdly, edge devices often have limited computational capabilities, making it difficult to efficiently support the training and updating of complex models. To address this bottleneck, an edge-cloud collaboration model can be adopted, where cloud computing power is used for offline model training and updating, and edge devices are only responsible for model inference and execution. In addition, lightweight alternative models, such as lightweight convolutional networks or small Transformers, can also be explored for model and hardware adaptation.

Fourthly, this framework involves multiple hyper-parameters across its different modules. Although grid search and empirical Settings are used, finding the optimal hyperparameter set can be time-consuming and may require domain expertise. Sensitivity analysis provides some insights, but more robust and automated tuning methods might be beneficial.

Finally, although the framework provides the best operation, the deep learning components may be black boxes, so it is difficult to fully understand why specific maintenance operations are recommended. Enhancing the interpretability of model decisions is valuable for gaining the trust and acceptance of maintenance engineers.

## 7. Conclusion

The MDP framework integrating FNO-DAE-GNN-PPO proposed in this paper systematically solves the key problems in predictive maintenance. Firstly, FNO is utilized for frequency-domain modeling to capture the complex dynamic patterns of device state evolution from high-dimensional time sensor data. Secondly, DAE is introduced for nonlinear dimensionality reduction to extract robust and compact state representations and enhance adaptability to noise and non-Gaussian feature distributions. Thirdly, considering the dependencies among devices, GNN models the structured context information by integrating the states of adjacent devices. Finally, the PPO algorithm is utilized to train and optimize the maintenance strategy based on these high-quality state embeddings and graph structure representations, in order to achieve the optimal control of long-term maintenance costs.

The experimental results prove the superiority of the FNO-DAE-GNN-PPO MDP framework. Compared with the baseline models (MDP + Q-learning, DQN-PCA MDP and Transformer-VAE MDP), our method achieves the lowest long-term average maintenance cost. This learning strategy also demonstrates a more detailed and progressive approach that enables maintenance decisions to be made under different equipment wear conditions. Although the proposed method has a higher computing time (92.19 seconds) compared with the baseline, it is acceptable for the strategic maintenance plan. The convergence analysis was conducted by tracking the KL divergence during the PPO training process, indicating that the strategy was successfully stabilized after the initial exploration stage, suggesting that the learning was effective. The ablation study further verified the contribution of each module (FNO, DAE, GNN, PPO), and no single component led to performance degradation, especially in terms of average cost or strategy consistency.

In conclusion, this study successfully developed and verified an integrated, multi-module deep reinforcement learning framework, which effectively addressed the challenges in time series modeling, state representation, and policy optimization for predictive maintenance. The MDP method based on FNO-DAE-GNN-PPO provides a powerful and efficient solution for optimizing maintenance decisions in complex manufacturing systems, and offers great potential for reducing costs and improving operational efficiency in industrial applications. Future work can explore further improvements to reduce computational complexity and study the adaptability of the framework to a wider range of industrial equipment and dynamic operating conditions.

## Acknowledgments

The authors would like to express our sincere appreciation to the editor and anonymous reviewers for their insightful comments, which greatly improved the quality of this paper.